\documentclass[10pt]{article}
\usepackage{arxiv}

\ifpdf \usepackage[pdftex]{graphicx} \pdfcompresslevel=9
\else \usepackage[dvips]{graphicx} \fi 

\usepackage{t1enc}
\usepackage{cite}

\title{ConvPoint: Continuous Convolutions for Point Cloud Processing}
\author{
  Alexandre Boulch\\
  ONERA, Universit\'{e} Paris-Saclay, FR-91123 Palaiseau, France\\
  valeo.ai, Paris, France
}

\subtitle{\color{red} Preprint. Accepted at Computer \& Graphics, see final version in journal issue.}

\usepackage{hyperref}
\usepackage{microtype}
\usepackage{subfigure}
\usepackage{booktabs}
\usepackage{amsfonts}       % blackboard math symbols
\usepackage{graphicx}
\usepackage{amsmath}
\usepackage{stmaryrd}
\newcolumntype{M}{@{\hspace{2pt}}>{\columncolor{white}[2pt][2pt]}c@{\hspace{2pt}}}

\begin{document}

\maketitle

\begin{abstract}
%%%
Point clouds are unstructured and unordered data, as opposed to images.
Thus, most machine learning approach developed for image cannot be directly transferred to point clouds.
In this paper, we propose a generalization of discrete convolutional neural networks (CNNs) in order to deal with point clouds by replacing discrete kernels by continuous ones.
This formulation is simple, allows arbitrary point cloud sizes and can easily be used for designing neural networks similarly to 2D CNNs.
We present experimental results with various architectures, highlighting the flexibility of the proposed approach.
We obtain competitive results compared to the state-of-the-art on shape classification, part segmentation and semantic segmentation for large-scale point clouds.
%%%%
\end{abstract}

%%%%%%%%%%%%%%%%%%%%%%%%%%%%%%%%%%%%%%%%%%%%%%%%%%%%%%%%%%%%%%%%%%%%%%%%%%%%%%%%%%%%%
%%%%%%%%%%%%%%%%%%%%%%%%%%%%%%%%%%%%%%%%%%%%%%%%%%%%%%%%%%%%%%%%%%%%%%%%%%%%%%%%%%%%%
%%%%%%%%%%%%%%%%%%%%%%%%%%%%%%%%%%%%%%%%%%%%%%%%%%%%%%%%%%%%%%%%%%%%%%%%%%%%%%%%%%%%%
% ██╗███╗   ██╗████████╗██████╗  ██████╗ 
% ██║████╗  ██║╚══██╔══╝██╔══██╗██╔═══██╗
% ██║██╔██╗ ██║   ██║   ██████╔╝██║   ██║
% ██║██║╚██╗██║   ██║   ██╔══██╗██║   ██║
% ██║██║ ╚████║   ██║   ██║  ██║╚██████╔╝
% ╚═╝╚═╝  ╚═══╝   ╚═╝   ╚═╝  ╚═╝ ╚═════
%%%%%%%%%%%%%%%%%%%%%%%%%%%%%%%%%%%%%%%%%%%%%%%%%%%%%%%%%%%%%%%%%%%%%%%%%%%%%%%%%%%%%
%%%%%%%%%%%%%%%%%%%%%%%%%%%%%%%%%%%%%%%%%%%%%%%%%%%%%%%%%%%%%%%%%%%%%%%%%%%%%%%%%%%%%
%%%%%%%%%%%%%%%%%%%%%%%%%%%%%%%%%%%%%%%%%%%%%%%%%%%%%%%%%%%%%%%%%%%%%%%%%%%%%%%%%%%%%

  \section{Introduction}

  \begin{figure}
    \begin{tabular}{cc}
      \includegraphics[width=0.48\linewidth]{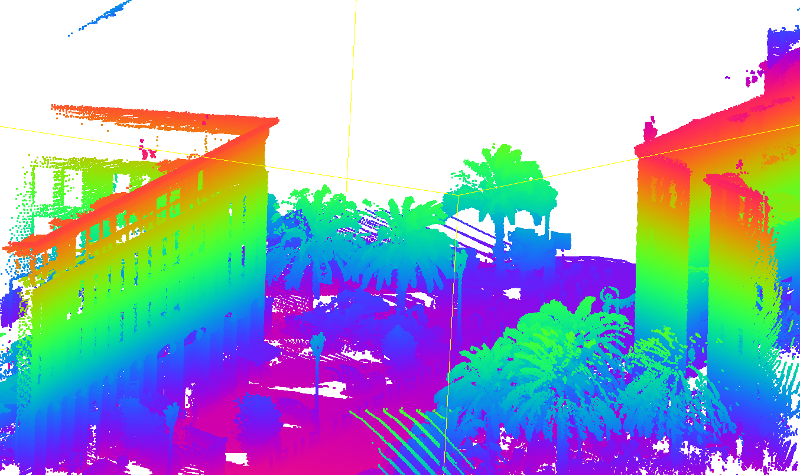} &
      \includegraphics[width=0.48\linewidth]{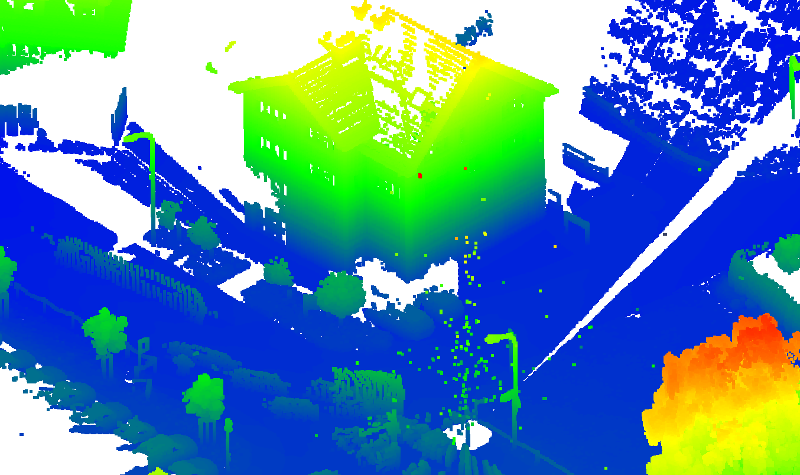} \\
      \includegraphics[width=0.48\linewidth]{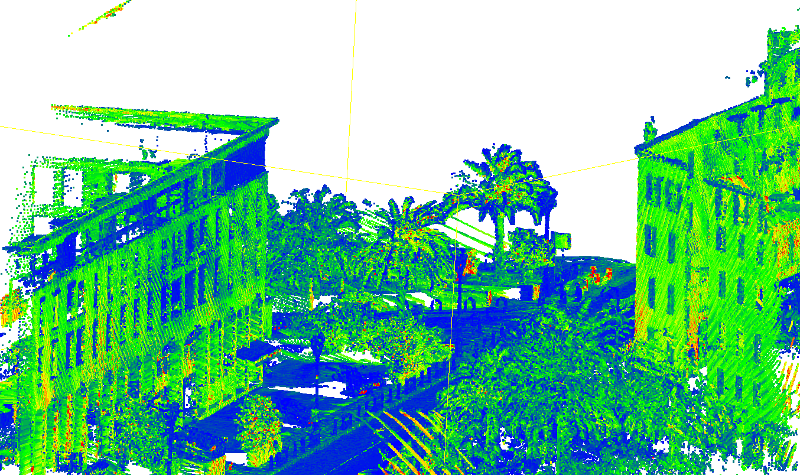} &
      \includegraphics[width=0.48\linewidth]{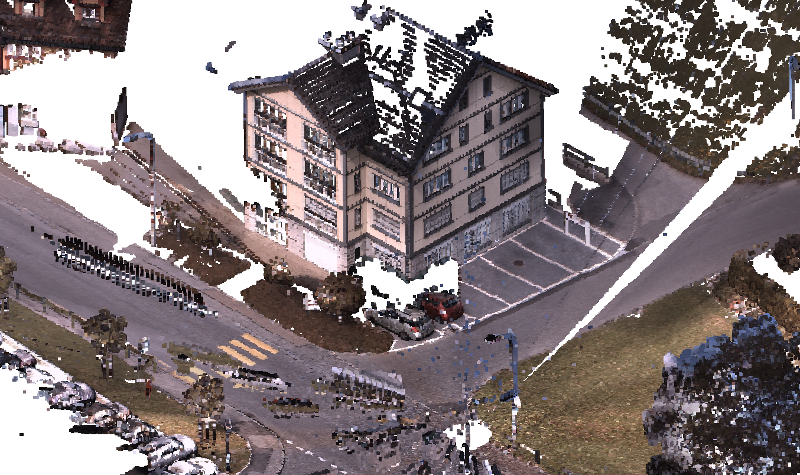} \\
      \includegraphics[width=0.48\linewidth]{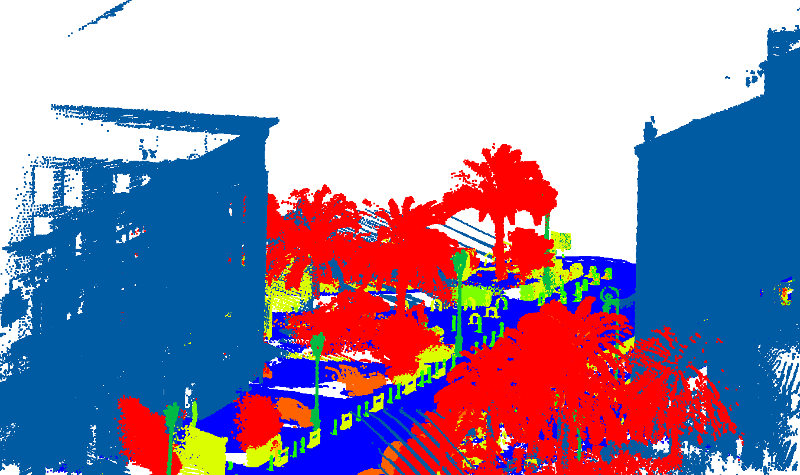} &
      \includegraphics[width=0.48\linewidth]{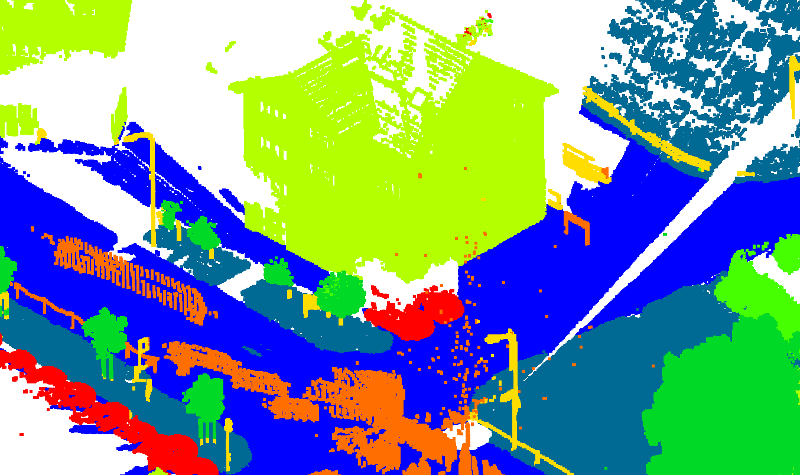} \\
      (a) NPM3D & (b) Semantic8 \\
    \end{tabular}
    \caption{Segmentation results on datasets (a) NPM3D and (b) Semantic8. Point cloud colored according height (top), laser intensity or RGB colors (middle) and predicted segmentation label (bottom).}
    \label{fig:visual_results}
  \end{figure}
 
  Point clouds are widely used in varied domains such as historical heritage preservation and autonomous driving.
  They can be either directly generated using active sensors such as LIDARs or RGB-Depth sensors, or an intermediary product of photogrammetry.
  
  These point sets are sparse samplings of the underlying surface of scenes or objects.
  With the exception of structured acquisitions (e.g., with LIDARs), point clouds are generally unordered and without spatial structure; they cannot be sampled on a regular grid and processed as image pixels.
  Moreover, the points may or may not hold colorimetric features. Thus, point-cloud dedicated processing must be able to deal only with the relative positions of the points.
  
  Due to these considerations, methods developed for image processing cannot be used directly on point clouds.
  In particular, convolutional neural networks (CNNs) have reached state-of-the-art performance in many image processing tasks.
  They use discrete convolutions which make extensive use the grid structure of the data, which does not exist with point clouds.
  
  Two common approaches to circumvent this problem are: first, to project the points into a space suitable for discrete convolutions, e.g., voxels; second, to reformulate the CNNs to take into account point clouds' unstructured nature.
  
  In this paper, we adopt the second approach and propose a generalization of CNNs for point clouds.
  The main contributions of this paper are two-fold.
    
  First, we introduce a continuous convolution formulation designed for unstructured data.
  The continuous convolution is a simple and straightforward extension of the discrete one.
    
  Second, we show that this continuous convolution can be used to build neural networks similarly to its image processing counterpart.
  We design neural networks using this convolution and a hierarchical data representation structure based on a search tree.
    
  We show that our framework, \textit{ConvPoint}, which extends our work presented in~\cite{boulch2019conv}, can be applied to various classification and segmentation tasks, including large scale indoor and outdoor semantic segmentation.
  For each task, we show that ConvPoint is competitive with the state-of-the-art.
  
  The paper is organized as follows: Section~\ref{sec_related} presents the related work, Section~\ref{sec_conv} describes the continuous convolutional formulation, Section~\ref{sec_tree} is dedicated to the spatial representation of the data, and Section~\ref{sec_layer} presents the convolution as a layer. Finally, Section~\ref{sec_expe} shows experiments on different datasets for classification and semantic segmentation of point clouds.

%%%%%%%%%%%%%%%%%%%%%%%%%%%%%%%%%%%%%%%%%%%%%%%%%%%%%%%%%%%%%%%%%%%%%%%%%%%%%%%%%%%%%
%%%%%%%%%%%%%%%%%%%%%%%%%%%%%%%%%%%%%%%%%%%%%%%%%%%%%%%%%%%%%%%%%%%%%%%%%%%%%%%%%%%%%
%%%%%%%%%%%%%%%%%%%%%%%%%%%%%%%%%%%%%%%%%%%%%%%%%%%%%%%%%%%%%%%%%%%%%%%%%%%%%%%%%%%%%
% ██████╗ ███████╗██╗      █████╗ ████████╗███████╗██████╗ 
% ██╔══██╗██╔════╝██║     ██╔══██╗╚══██╔══╝██╔════╝██╔══██╗
% ██████╔╝█████╗  ██║     ███████║   ██║   █████╗  ██║  ██║
% ██╔══██╗██╔══╝  ██║     ██╔══██║   ██║   ██╔══╝  ██║  ██║
% ██║  ██║███████╗███████╗██║  ██║   ██║   ███████╗██████╔╝
% ╚═╝  ╚═╝╚══════╝╚══════╝╚═╝  ╚═╝   ╚═╝   ╚══════╝╚═════╝
%%%%%%%%%%%%%%%%%%%%%%%%%%%%%%%%%%%%%%%%%%%%%%%%%%%%%%%%%%%%%%%%%%%%%%%%%%%%%%%%%%%%%
%%%%%%%%%%%%%%%%%%%%%%%%%%%%%%%%%%%%%%%%%%%%%%%%%%%%%%%%%%%%%%%%%%%%%%%%%%%%%%%%%%%%%
%%%%%%%%%%%%%%%%%%%%%%%%%%%%%%%%%%%%%%%%%%%%%%%%%%%%%%%%%%%%%%%%%%%%%%%%%%%%%%%%%%%%%

\section{Related work}
\label{sec_related}

Point cloud processing is a widely discussed topic.
We focus here on machine learning techniques for point cloud classification or local attribute estimation.

Most of the early methods use handcrafted features defined using a point and its neighborhood, or a local estimate of the underlying surface around the point.
These features describe specific properties of the shape and are designed to be invariant to rigid or non rigid transformations of the shape \cite{johnson1999using, aubry2011wave, bronstein2010scale, ling2007shape}.
Then, classical machine learning algorithms are trained using these descriptors.

In the last years, the release of large annotated point cloud databases has allowed the development of deep neural network methods that can learn both descriptors and decision functions.

%voxels
The direct adaptation of CNNs developed for image processing is to use 3D convolutions.
Methods like \cite{wu20153d,maturana2015voxnet} apply 3D convolutions on a voxel grid.
Even though recent hardware advances enable the use of these networks on larger scenes, they are still time consuming and require a relatively low voxel resolution, which may result in a loss of information and undesirable bias due to grid axis alignment.
In order to avoid these drawbacks, \cite{SubmanifoldSparseConvNet, graham20183d} use sparse convolutional kernels and \cite{li2016fpnn, wang2015voting} scan the 3D space to focus computation where objects are located.

% Multiview
A second class of algorithms avoids 3D convolutions by creating 2D representations of the shape, applying 2D CNNs and projecting the results back to 3D.
This approach has been used for object classification \cite{su2015multi}, jointly with voxels \cite{qi2016volumetric}, and for semantic segmentation \cite{boulch2017snapnet}.
One of the main issues of using multi-view frameworks is to design an efficient and robust strategy for choosing viewpoints.

% graph
The previous methods are based on 2D or 3D CNNs, creating the need to structure the data (3D grid or 2D image) in order to process it.
Recently, work has focused on developing deep learning architectures for non-Euclidean data.
This work, referred to as \textit{geometric deep learning}~\cite{bronstein2017geometric}, operates on manifolds, graphs, or directly on point clouds.

Neural networks on graphs were pioneered in~\cite{scarselli2008graph}. Since then, several methods have been developed using the spectral domain from the graph Laplacian~\cite{shuman2013emerging,bruna2013spectral,yi2017syncspeccnn} or polynomials \cite{defferrard2016convolutional} as well as gated recurrent units for message passing~\cite{li2015gated,landrieu2018large}.

The first extension of CNNs to manifolds is Geodesic CNN~\cite{masci2015geodesic}, which redefines usual layers such as convolution or max pooling. It is applied to local mesh patches in a polar representation to learn invariant shape features for shape correspondences.
In~\cite{boscaini2016learning}, the representation is improved with anisotropic diffusion kernels and the resulting method is not bound to triangular meshes anymore.
More recently, MoNet~\cite{monti2017geometric} offers a unified formulation of CNNs on manifolds which included \cite{masci2015geodesic, atwood2016diffusion, boscaini2016learning} and reaches state-of-the-art performance on shape correspondence benchmarks.

These methods have proved to be very efficient but they usually operate on graphs or meshes for surface analysis.
However, building a mesh from raw point clouds is a very difficult task and requires in practice priors regarding the surface to be reconstructed~\cite{berger2017survey}.

A fourth class of machine learning approaches processes directly the raw point clouds; the method proposed in this paper belongs to this category.

One of the recent breakthroughs in dealing with unstructured data is PointNet~\cite{qi2017pointnet}.
The key idea is to construct a transfer function invariant by permutation of the inputs features, obtained by using the order invariant \textit{max pooling} function.
The coordinates of the points are given as input and geometric operations (affine transformations) are obtained with small auxiliary networks.
However, it fails to capture local structures. Its improvement, PointNet++ \cite{qi2017pointnet++}, uses a cascade of PointNet networks from local scale to global scale.
\\

Convolutional neural layers are widely used in machine learning for image processing and more generally for data sampled on a regular grid (such as 3D voxels).
However, the use of CNNs when data is missing or not sampled on a regular grid is not straightforward.
Several works have studied the generalization of such powerful schemes to such cases.

%Deformable convolutions SplatNet
To avoid problems linked to discrete kernels and sparse data.
\cite{dai2017deformable} introduces deformable convolutional kernels able to adapt to the recognition task.
In \cite{su2018splatnet}, the authors adapt~\cite{dai2017deformable} to deal with point clouds.
The input signal is interpolated on the convolutional kernel, the convolution is applied, and the output is interpolated back to input shape.
The kernel elements' locations are optimized like in \cite{dai2017deformable} and the input points are weighted according to their distance to kernel elements, like in \cite{su2018splatnet}.
However, unlike in \cite{su2018splatnet}, the our approach is not dependent of a convolution kernel designed on a grid.

In \cite{NIPS2018_7362}, a $\chi$-transform is applied on the point coordinates to create geometrical features to be combined with the input point features.
This is a major difference relative to our approach: as in \cite{qi2017pointnet++}, \cite{NIPS2018_7362} makes use of the input geometry as features. 
We want the geometry to behave as in discrete convolutions, i.e. weighting the relation between kernel and input. In other words, the geometric aspects are defined in the network structure (convolution strides, pooling layers) and point coordinates (e.g. pixels indices for images) are not an input of the network.

In \cite{wang2018deep}, the authors propose a convolution layer defined as a weighted sum of the input features, the weights being computed by a multi-layer perceptron (MLP). Our approach has points in common with~\cite{wang2018deep} both in concept and implementation, but differs in two ways. 
First, our approach computes a dense weighting function that takes into account the whole kernel.
Second, as in~\cite{thomas2019KPConv}, the derived kernel is an explicit set of points associated with weights.
However, whereas~\cite{thomas2019KPConv} uses an explicit RBF Gaussian function to correlate input and kernel, we propose to learn this kernel to input relation function with a multi-layer perceptron (MLP).

\section{Convolution for point processing}
\label{sec_conv}

We build our continuous convolutional layer by adapting the discrete convolution formulation used for grid-sampled data such as images.

\textit{Notations.} In the following sections, $d$ is the dimension of the spatial domain (e.g., $3$ for 3D point clouds) and $n$ is the dimension of the features domain (i.e., the dimension of the input features of the convolutional layer).
 $\llbracket a, b\rrbracket$ is an integer sequence from $a$ to $b$ with step $1$. The cardinality of a set $S$ is denoted by $|S|$.

\subsection{Discrete convolutions}

Let $K=\{\mathbf{w}_i, \in \mathbb{R}^n, i \in \llbracket 1,|K|\rrbracket \}$ be the kernel and  $X=\{\mathbf{x}_i, \in \mathbb{R}^n, i \in \llbracket 1, |X|\rrbracket\}$ be the input.
In discrete convolutions, $K$ and $X$ have the same cardinality.

We first consider the case where the elements of $K$ and $X$ range over the same locations (e.g., grid coordinates on an image).

Given a bias $\beta$, the output $y$ is:
\begin{equation}
  y = \beta + \sum_{i=1}^{|K|}  \mathbf{w}_i \mathbf{x}_i =
  \beta + \sum_{i=1}^{|K|} \sum_{j=1}^{|X|}  \mathbf{w}_i \mathbf{x}_j \mathbf{1}(i,j)
\end{equation}
where $\mathbf{1}(.,.)$ is the indicator function such that $\mathbf{1}(a,b)=1$ if $a=b$ and $0$ otherwise.
This expresses a one-to-one relation between kernel elements and input elements.

We now reformulate the previous equation by making explicit the underlying order of sets $K$ and $X$.
Let's consider $K = \{(\mathbf{c}, \mathbf{w})\}$ (resp. $X = \{(\mathbf{p}, \mathbf{x})\}$) where $\mathbf{c}_i \in \mathbb{R}^d$ (resp. $\mathbf{p}_j \in \mathbb{R}^d$) is the spatial location of the kernel element $i$ (resp. $j$ is the spatial location of the $j$-th point in the input).
For convolutions on images, $\mathbf{c}$ and $\mathbf{p}$ denote pixel coordinates in the considered patch.
The output is then:
\begin{equation}
  y = \beta + \sum_{j=1}^{|X|} \sum_{i=1}^{|K|}  \mathbf{w}_i \mathbf{x}_j \mathbf{1}(\mathbf{c}_i, \mathbf{p}_j)
  \label{eq:discrete}
\end{equation}

\subsection{Convolutions on point sets}

In this study, we consider that elements of $X$ may have any spatial locations, i.e., we do not assume a grid or other structure on $X$.
In practice, using equation~\eqref{eq:discrete} on such an $X$ would result in $\mathbf{1}(\mathbf{c}, \mathbf{p})$ to be zero (almost) all the time as the probability for a random $\mathbf{p}$ to match exactly $\mathbf{c}$ is zero.
Thus, using the indicator function is too restrictive to be useful when processing point clouds.

We need a more general function $\phi$ to establish a relation between the points $\{\mathbf{p}\}$ and the kernel elements $\{\mathbf{c}\}$.
We define $\phi$ as a geometrical weighting function that distributes the input $\mathbf{x}$ onto the kernel.

$\phi$ must be invariant to permutations of the input points (it is necessary since point clouds are unordered).
This is achieved if $\phi$ is independently applied on each point $\mathbf{p}$.
It can also be function of the set kernel points, i.e., $\{\mathbf{c}\}$.

Moreover, to be invariant to a global translation applied on both the kernel and the input cloud, we used relative coordinates with respect to kernel elements, i.e., we apply the $\phi$ function on $\{\mathbf{p}_j - \mathbf{c}_i\}, i\in  \llbracket 1, |K| \rrbracket$, the set of relative positions of $\mathbf{p}_j$ relatively to kernel elements.

%Then, each kernel element may have its own function $\phi_i$.

%We can consider a different $\phi_i$ instance of the function for each kernel element:
%
The $\phi$ function is then a function such that:
\begin{align}
  \phi\colon \mathbb{R}^d \times (\mathbb{R}^d)^{|K|} &\longrightarrow \mathbb{R}\\
  (\mathbf{p}, \{\mathbf{c}\} ) &\mapsto 
    \phi(\{\mathbf{p}-\mathbf{c}\})
\end{align}

Finally, the convolution operation for point sets is defined by:
\begin{equation}
\label{eq:conv}
    y = \beta + \frac{1}{|X|}\sum_{j=1}^{|X|} \sum_{i=1}^{|K|}  \mathbf{w}_i \, \mathbf{x}_j \, \phi(\{\mathbf{p}_j-\mathbf{c}\})
\end{equation}
where we added a normalization according to the input set size for robustness to variation in input size.

In this formulation, the spatial domain and the feature domain are mixed similarly to the discrete formulation.
Unlike PointNet~\cite{qi2017pointnet} and PointNet++~\cite{qi2017pointnet++}, spatial coordinates are not input features.
This formulation is closer to~\cite{wang2018deep} where the authors estimate directly the weights of the input features, i.e., the product $\mathbf{w}_i \, \phi(\{\mathbf{p}_j-\mathbf{c}\})$, mixing estimation in the feature space and the geometrical space. 

%%%% FIG Layer
\begin{figure}
  \centering
  \includegraphics[width=0.65\linewidth]{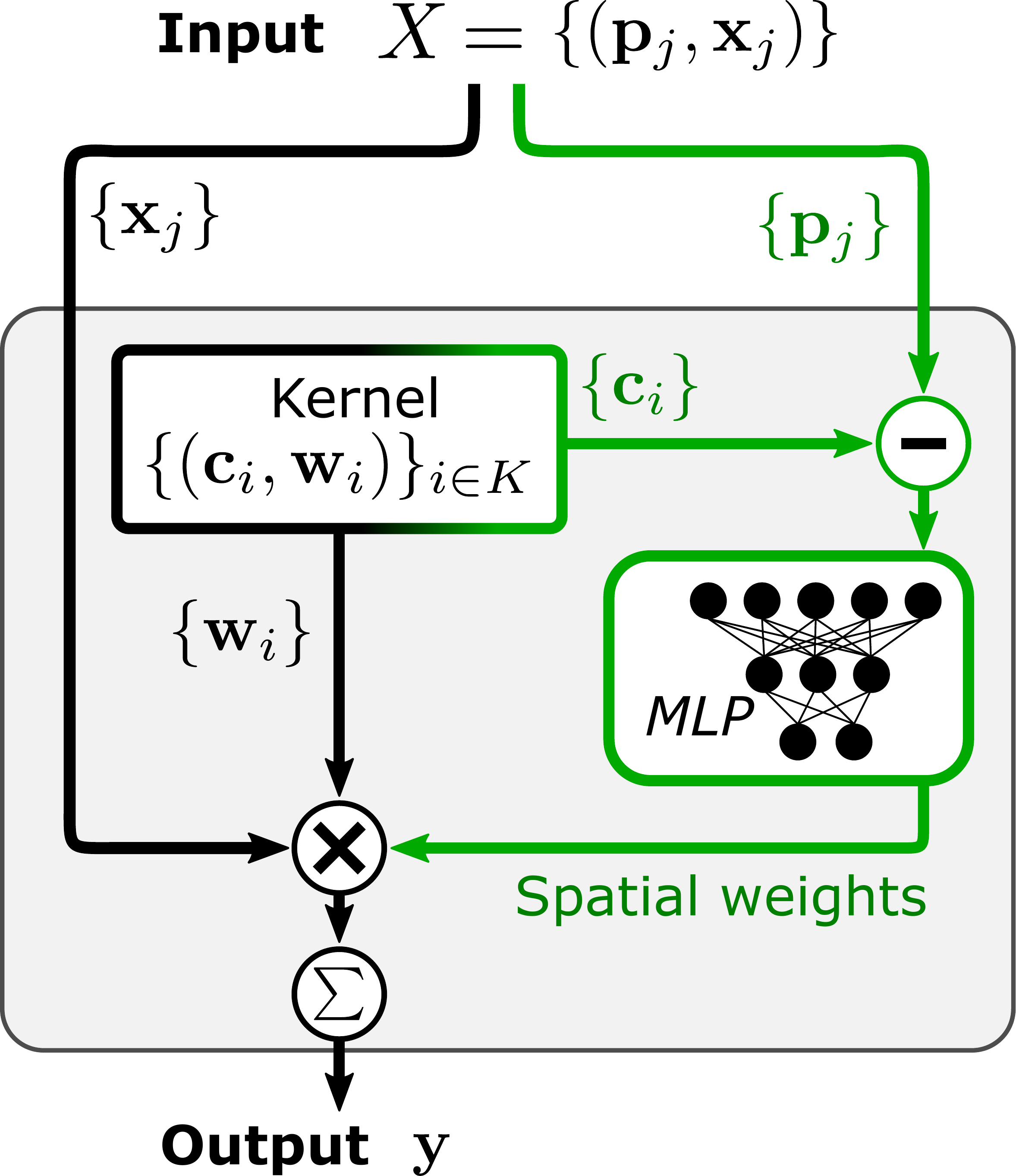}
  \caption{
  Convolution operation. Black arrows and boxes with black frames are features operations, green ones are spatial operations.
  }
  \label{fig:layer}
\end{figure}

\subsection{Weighting function}

In practice designing by hand such a $\phi$ function is not easy.
Intuitively, it would decrease with the norm of $\mathbf{p} - \mathbf{c}$.
As in \cite{thomas2019KPConv}, Gaussian functions could be used for that, but it would require handcrafted parameters which are difficult to tune.
Instead, we choose to learn this function with a simple MLP.
Such an approach does not make any specific assumption about the behavior of the function $\phi$.

\subsection{Convolution}

The convolution operation is illustrated in Fig.~\ref{fig:layer}.
For a kernel $\{(\mathbf{c}, \mathbf{w})\}$, the spatial part $\{\mathbf{c}\}$ and feature part $\{\mathbf{w}\}$ are processed separately.
The black arrows and boxes are for operations in the features space, green ones are for operations in the point cloud space.
Please note that removing the green operations corresponds to the discrete convolution.

Moreover, as in the case of discrete convolutions, our convolutional layer is made of several kernels (corresponding to the number of output channel). The output of the convolution is thus a vector $\mathbf{y}$.

\subsection{Parameters and optimization on kernel element location}

To set the location of the kernel elements, we randomly sample the locations $\mathbf{c} \in \{\mathbf{c}\}$ in the unit sphere and consider them as parameters.
As $\phi$ is differentiable with respect to its input, $\{\mathbf{c}\}$ can be optimized.

At training time, both parameters $\{\mathbf{w}\}$ and $\{ \mathbf{c} \}$ of $K$ as well as the MLP parameters are optimized using gradient descent.

\subsection{Properties}

\textit{Permutation invariance.}~~~
As stated in \cite{qi2017pointnet}, operators on point clouds must be invariant unser the permutation of points.
In the general case, the points are not ordered.
In our formulation, $y$ is a sum over the input points.
Thus, permutations of the points have no effect on the results.

\textit{Translation invariance.}~~~
As the geometric relations between the points and the kernel elements are relative, i.e., ($\{\mathbf{p}-\mathbf{c}\}$), applying a global translation to the point cloud does not change the results.

\textit{Insensitivity to the scale of the input point cloud.}~~~
Many point clouds, such as photogrammetric point clouds, have no metric scale information. Moreover the scale may vary from one point cloud to another inside a dataset.
In order to make the convolution robust to the input scale, the input geometric points $\{\mathbf{p}\}$ are normalized to fit in the unit ball.

\textit{Reduced sensibility to the size of input point cloud.}~~~
Dividing $\mathbf{y}$ by $|X|$ makes the output less sensitive to input size.
% For instance., using "$X + X $" as input does not change the result.
For instance, using "$2 X$" as input does not change the result.

\begin{figure*}
\includegraphics[width=\linewidth]{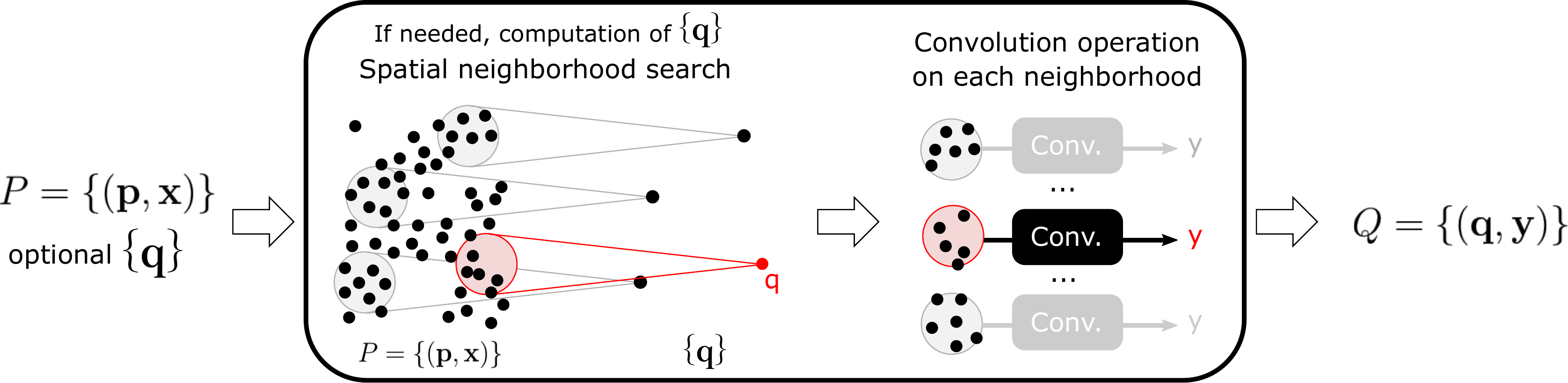}
\caption{
    %Convolutional layer.
    Convolutional layer, composed of two steps: the spatial structure computation (selecting $\{q\}$ if needed by computing the local neighborhoods of each $q$) and the convolution operation itself on each neighborhood, as defined in figure~\ref{fig:layer}.
    }
\label{fig:layer_conv}
\end{figure*}

\section{Hierarchical representation and neighborhood computation}
\label{sec_tree}

Let $P$ be an input point cloud.
The convolution as described in the previous section is a local operation on a subset $X$ of $P$.
It operates a projection of $P$ on an output point cloud $\{\mathbf{q}\}$ and computes the features $\{ \mathbf{y} \}$ associated with each point of $Q=\{\mathbf{q}, \mathbf{y}\}$.

Depending on the cardinality of $Q$, we have three possible behaviors for the convolution layer:

(a) {$|P|=|Q|$}. In this case the cardinality of the output is the same as the intput.
In the discrete framework it corresponds to the convolution without stride.

(b) {$|P|>|Q|$}. This includes the particular case of $\{\mathbf{p}\} \supset \{\mathbf{q}\}$.
The convolution operates a spatial size reduction.
The parallel with the discrete framework is a convolution with stride bigger than one.

(c) {$|P|<|Q|$}. This includes the particular case of $\{\mathbf{p}\} \subset \{\mathbf{q}\}$.
The convolutional layer produces an upsampled version of the input. It is similar to the up-convolutional layer in discrete pipelines.

\paragraph{Computation of $\{\mathbf{q}\}$}
$\{\mathbf{q}\}$ can either be given as an input, or computed from $\{\mathbf{p}\}$.
For the second case, a possible strategy is to randomly pick points in $\{\mathbf{p}\}$ to generate $\{\mathbf{q}\}$.
However, it is possible to pick several times the same point and some points of $\{\mathbf{p}\}$ may not be in the neighborhoods of the points of $\{\mathbf{q}\}$.
An alternative is proposed in~\cite{qi2017pointnet++} using a furthest-point sampling strategy. This is very efficient and ensures a spatially uniform sampling, however it requires to compute all mutual distances in the point cloud.

We propose an intermediate solution.
For each point, we memorize how many times it has been selected.
We pick the next point in the set of points with the lower number of selection.
Each time a point $\mathbf{q}$ is selected, its score is increased by $100$.
The score of the points in its neighborhood are increased by $1$.
The points of $\{\mathbf{q}\}$ are iteratively picked until the specified number of points is reached.
Using a higher score for the points in $\{\mathbf{q}\}$ ensures that they will not be chosen anymore, except if all points have been selected once.

\paragraph{Neighborhood computation} All $k$-nearest neighbor search are computed using a kd-tree built with $\{\mathbf{p}\}$.

\section{Convolutional layer}
\label{sec_layer}

The convolutional layer is presented on Fig.~\ref{fig:layer_conv}. It is composed of the two previously described operations (point selection and the convolution itself).
The inputs are $P$ and optionally $\{\mathbf{q}\}$.
If $\{\mathbf{q}\}$ is not provided, it is selected as a subset of $P$ following the procedure described in the previous section.
First, for each point of $\{\mathbf{q}\}$, local neighborhoods in $P$ are computed using a k-d tree. 
Then, for each of these subsets, we apply the convolution operation, creating the output features.
Finally, the output $Q$ is the union of the pairs $\{ (\mathbf{q}, \mathbf{y})\}$.

\paragraph*{Parameters}
The parameters of the convolutional layers are very similar to discrete convolution parameters in the most deep learning frameworks.
\begin{itemize}\setlength\itemsep{0.em}
  \item[-] \textbf{Number of output channels} ($C$): it is the number of convolutional kernels used in the layer. It defines the output feature dimension, i.e., the dimension of $\mathbf{y}$.
  \item[-] \textbf{Size of the output point cloud} ($|Q|$): it is the number of points that are passed to the next layer.
  \item[-] \textbf{Kernel size} ($|K|$): it is the number of kernel elements used for the convolution computation.
  \item[-] \textbf{Neighborhood size} ($k$): it is the number of points in $\{\mathbf{p}\}$ to consider for each point in $\{\mathbf{q}\}$.
\end{itemize}
\section{Experiments}
\label{sec_expe}

The following section is dedicated to experiments and comparison with state-of-the-art methods.
As the spatial structure generation (selection of the output point cloud for each layer) is a stochastic process, two runs through the network may lead to different outputs.
In the following, we aggregate the results of multiple runs by averaging the outputs.
The number of runs is then referred to as the number of spatial samplings. In the folowing tables, it correspond to the number between parentheses (for classification and part segmentation).
The influence of this number is discussed in section~\ref{sec:spatial_sampling}.

\subsection{Classification}

  The first experiments are dedicated to points cloud classification.
  We experimented on both 2D and 3D point cloud datasets.

  %%%% FIG Network classif
  \begin{figure}
    \centering
    \includegraphics[width=\linewidth]{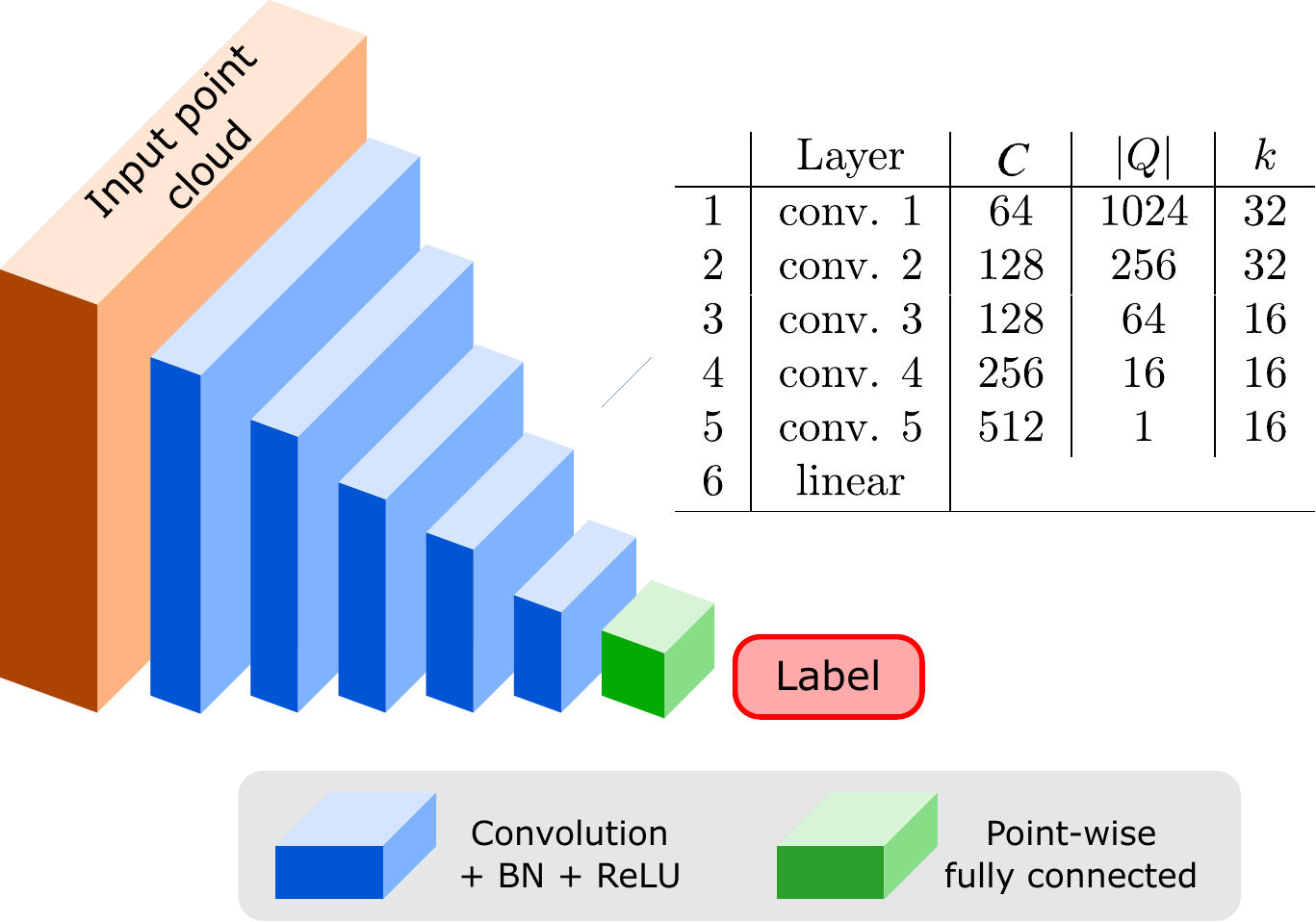}
    \caption{
        Classification network.
        It is composed of five convolution layer (blue blocks) with progressive point cloud size reduction ($|Q|$) and a final fully connected layer (green block). For all layers $|K|=16$.
    }
    \label{fig:net_classif}
  \end{figure}

  \paragraph*{Network}

  The network is described in Fig.~\ref{fig:net_classif}.
  It is composed of five convolutions that progressively reduce the point cloud size to one single point, while increasing the number of channels.
  The features associated to this point are the inputs of a linear layer.
  This architecture is very similar to the ones that can be used for image processing (e.g., LeNet~\cite{lecun1998gradient}).

  \paragraph*{2D classification: MNIST}

  The 2D experiment is done on the MNIST dataset.
  This is a dataset for the classification of gray scale handwritten digits.
  The point cloud $P=\{(\mathbf{p},\mathbf{x})\}$ is built from the images, using pixel coordinates as point coordinates and, thus, is sampled on a grid.
  We build two variants of the dataset: first, point clouds are built with the whole image and the features associated with each point is the grey level ($\{\mathbf{x}\}=\{\text{grey level}\}$); second, only the black points are considered ($\{\mathbf{x}\}=\{1\}$).

  Results are presented in table~\ref{tab:classif}(a).
  We compare with both image CNNs (LeNet~\cite{lecun1998gradient} and Network in Network~\cite{lin2013network}) and point-based methods (PointNet++~\cite{qi2017pointnet++} and PointCNN~\cite{NIPS2018_7362}).
  Scores, averaged over 16 spatial samplings, are competitive with other methods.
  More interestingly, we do not observe a great difference between the two variants (grayscale points or black points only). 
  In the \textit{Gray levels} experiment (whole image), the framework is able to learn from the color value only as the points do not hold shape-related information. On the contrary, in the \textit{Black points only}, it learns from geometry only, which is a common case for point cloud.
  
  % In the first case, it shows that the framework is able to learn only from color value, which is the image case. In the second case, it learns from geometry only, which is a common case for point cloud.
  
  \paragraph*{3D classification: ModelNet40}
  We also experimented on 3D classification on the ModelNet40 dataset.
  This dataset is a set of meshes from 40 various classes (planes, cars, chairs, tables...).
  We generated point clouds by randomly sampling points on the triangular faces of the meshes.
  In our experiments, we use an input size of either 1024 or 2048 points for training.
  Table~\ref{tab:classif}(b) presents the results.
  % Scores corresponding to our method are averaged over ten trainings, we also display scores for the best of these ten runs.
  As for 2D classification, we are competitive with the state of the art concerning point-based classification approaches.

  %%%% TABLE Classif
  \begin{table}

    \caption{Shape classification. Overall accuracy (OA \%) and class average accuracy (AA, \%).}
    \label{tab:classif}
    \centering
    \small
    \begin{tabular}{cc}~\\
      (a) MNIST & (b) ModelNet 40 \\ ~\\
      \begin{tabular}[t]{@{}l@{\hspace{0.1cm}}|@{\hspace{0.1cm}}c@{}}
         Methods & OA \\
         \hline\hline
         \multicolumn{2}{c}{\textit{Image-based methods}} \\
         LeNet~\cite{lecun1998gradient} & 99.20 \\
         NiN~\cite{lin2013network} & 99.53 \\
         \hline\hline
         \multicolumn{2}{c}{\textit{Point-based methods}}\\
         PointNet++~\cite{qi2017pointnet++} & 99.49 \\
         PointCNN~\cite{NIPS2018_7362} & 99.54 \\
        %  ~\\
        %  ~\\
         \hline
        % Ours - ConvPoint\\
        Ours - Gray levels (16)& \textbf{99.62} \\
        Ours - Black points (16)& 99.49
      \end{tabular}&
      \begin{tabular}[t]{@{}l@{\hspace{0.1cm}}|@{\hspace{0.1cm}}c@{\hspace{0.2cm}}c@{}}
        Methods & OA & AA\\
        \hline\hline
        \multicolumn{2}{c}{\textit{Mesh or voxels}}\\
        Subvolume~\cite{qi2016volumetric}   & 89.2 \\
        MVCNN~\cite{su2015multi}            & 90.1 \\
        \hline\hline
        \multicolumn{2}{c}{\textit{Points}} \\
        DGCNN~\cite{wang2018dynamdynamic}        & 92.2 & \textbf{90.2} \\
        PointNet~\cite{qi2017pointnet}      & 89.2 & 86.2 \\ 
        PointNet++~\cite{qi2017pointnet++}  & 90.7 \\
        PointCNN~\cite{NIPS2018_7362}       & 92.2 & 88.1\\
        KPConv~\cite{thomas2019KPConv}      & \textbf{92.9} \\
        \hline
        % \textit{Ours}& \\
        % Ours - ConvPoint& \\
        % 1024 pts (16) avg. & 91.4 & 87.9 \\
        % \quad\quad\quad\quad\quad~ best & (91.8) & (88.5) \\
        Ours - 1024 pts (16) & 91.8 & 88.5 \\
        % 2048 pts (16) avg. & 92.1 & 89.3 \\
        % \quad\quad\quad\quad\quad~ best & (\textbf{92.5}) & (89.6) \\
        Ours - 2048 pts (16) & 92.5 & 89.6 \\
      \end{tabular}
    \end{tabular}
  \end{table}

\subsection{Segmentation}

  %%%% FIG network seg
  \begin{figure*}
    \centering
    \begin{tabular}{cc}
    \includegraphics[width=0.6\linewidth]{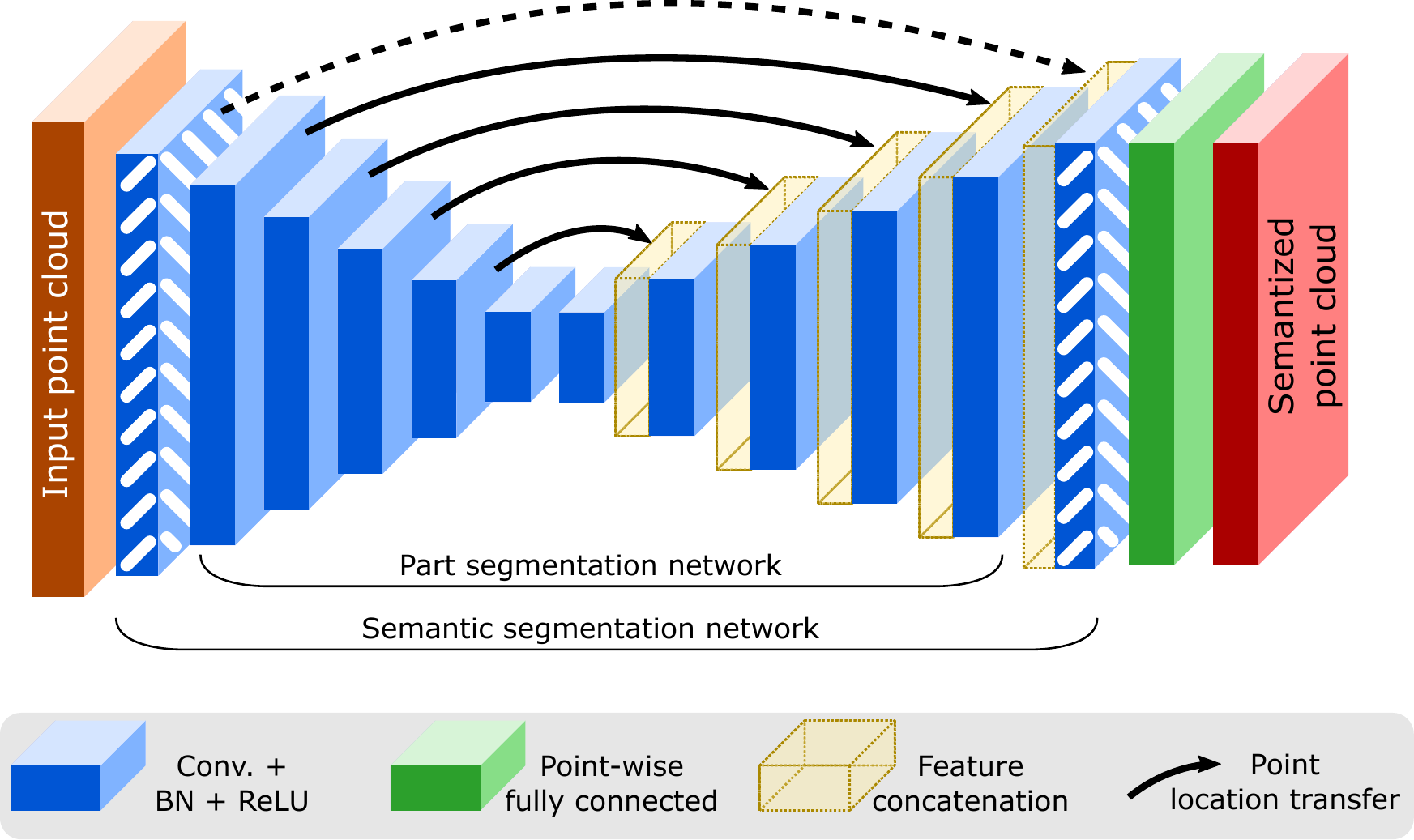} &
      \small
      \begin{tabular}[b]{c|c|c|c|c}
        &Layer & $C$ & $|Q|$ & $k$ \\
        \hline
         (0) & conv. 0     &  64 & $|P|$& 16\\
         (1) & conv. 1     &  64 & 2048 & 16\\
          2  & conv. 2     &  64 & 1024 & 16\\
          3  & conv. 3     &  64 &  256 & 16\\
          4  & conv. 4     & 128 &   64 &  8\\
          5  & conv. 5     & 128 &   16 &  8\\
          6  & conv. 6     & 128 &    8 &  4\\
          7  & (de)conv. 5 & 128 &   16 &  4\\
          8  & (de)conv. 4 & 128 &   64 &  4\\
          9  & (de)conv. 3 &  64 &  256 &  4\\
         10  & (de)conv. 2 &  64 & 1024 &  4\\
         11  & (de)conv. 1 &  64 & 2048 &  8\\
        (12) & (de)conv. 0 &  64 & $|P|$&  8\\
         13 & linear        \\
        \hline
        \multicolumn{1}{c}{~}\\
        \multicolumn{1}{c}{~}
      \end{tabular}
    \end{tabular}

    \textit{Note: striped convolution (layer 0, 1 and 12) are only in semantic segmentation network.}

    \caption{Segmentation networks.
        It is made of an encoder (progressive reduction of the point cloud size) followed by a decoder (to get back to the original point cloud size). Skip connections (black arrows) allow information to flow directly from encoder to decoder: they provide point locations for decoder's convolutions and encoder's features are concatenated with decoder's ones at corresponding scale.
    }
    \label{fig:net_seg}
  \end{figure*}

  \subsubsection{Segmentation network}

  The segmentation network is presented on Fig.~\ref{fig:net_seg}.
  It has an encoder-decoder structure, similar to U-Net, a stack of convolutions that reduces the cardinality of the point cloud as an encoder and a symmetrical stack of convolution as a decoder with skip connections.
  In the decoder, the points used for upsampling are the same as the points in the encoder at the corresponding layer.
  Following the U-Net architerture, the features from the encoder and from the decoder are concatenated at the input of the convolutional layers of the decoder.
  Finally, the last layer is a point-wise linear layer used to generate an output dimension corresponding to the number of classes.
  
  The network comes in two variants, i.e., with two different numbers of layers.
  The part segmentation network (plain colors, in Fig.~\ref{fig:net_seg}) is used with ShapeNet for part segmentation.
  The second network, used for large-scale segmentation, is the same network with three added convolutions (hatched layers in Fig.~\ref{fig:net_seg}).
  It is a larger network; its only purpose is to deal with larger input point clouds.

  For both versions of the network, we add a dropout layer between the last convolution and the linear layer.
  At training time, the probability of an element to be set to zero is 0.5.

  \subsubsection{Part segmentation}

  Given a point cloud, the part segmentation task is to recognize the different constitutive parts of the shape.
  In practice, it is a semantic segmentation at shape level.
  We use the Shapenet~\cite{yi2016scalable} dataset.
  It is composed of 16680 models belonging to 16 shape categories and split in train/test sets.
  Each category is annotated with 2 to 6 part labels, totalling 50 part classes.
  % The total is 50 part classes.
  As in~\cite{NIPS2018_7362}, we consider the part annotation problem as a 50-class semantic segmentation problem.
  We use a cross-entropy loss for each category.
  The scores are computed at shape level.

  We use the semantic segmentation network from Fig.~\ref{fig:net_seg}.
  The provided point clouds have various sizes, we randomly select 2500 points (possibly with duplication if the point cloud size is lower than 2500) and predict the labels for each input point.
  The points do not have color features so we set the input features to one (this is the same as for the MNIST experiment with black points only).
  As all points may not have been selected for labeling, the class of an unlabeled point is given by the label of its nearest neighbor.

  The results are presented in table~\ref{tab:shapenet}.
  The scores are the mean class intersection over union (mcIoU) and instance average intersection over union (mIoU).
  Similarly to classification, we aggregate the scores of multiple runs through the network.
  Our framework is also competitive with the state-of-the-art methods as we rank among the top five methods for both mcIoU and mIou.

  \begin{table}[t]
    \caption{ShapeNet}
    \label{tab:shapenet}
    \begin{minipage}{0.68\linewidth}
    \centering
    \small
    \begin{tabular}{l|cc}
        Method & mcIoU & mIoU \\
        \hline
        SyncSpecCNN~\cite{yi2017syncspeccnn}  & 82.0 & 84.7 \\
        Pd-Network~\cite{klokov2017escape}    & 82.7 & 85.5 \\
        3DmFV-Net~\cite{ben20173d}            & 81.0 & 84.3 \\
        PointNet~\cite{qi2017pointnet}        & 80.4 & 83.7 \\
        PointNet++~\cite{qi2017pointnet++}    & 81.9 & 85.1 \\
        SubSparseCN~\cite{graham20183d}              & 83.3 & 86.0 \\
        SPLATNet~\cite{su2018splatnet}        & 83.7 & 85.4 \\
        SpiderCNN~\cite{xu2018spidercnn}      & 81.7 & 85.3 \\
        SO-Net~\cite{li2018so}                & 81.0 & 84.9 \\
        PCNN~\cite{atzmon2018point}           & 81.8 & 85.1 \\
        KCNet~\cite{shen2018mining}           & 82.2 & 83.7 \\
        SpecGCN~\cite{wang2018local}          &    - & 85.4 \\
        RSNet~\cite{huang2018recurrent}       & 81.4 & 84.9 \\
        DGCNN~\cite{wang2018dynamdynamic}     & 82.3 & 85.1 \\
        SGPN~\cite{wang2018sgpn}              & 82.8 & 85.8 \\
        PointCNN~\cite{NIPS2018_7362}         & 84.6 & 86.1 \\
        KPConv~\cite{thomas2019KPConv}        & \textbf{85.1} & \textbf{86.4} \\

        \hline
        % ConvPoint \\
        % NetShapeNet (W False)  & 82.8 & 85.5 \\
        % NetShapeNet (W True)   & 83.5 & 84.2 \\
        % NetShapeNet3 (W False) & 83.1 & 85.9 \\
        % NetShapeNet3 (W True)  & 83.6 & 85.1 \\
        % NetShapeNet2 64 planes (W False)
        % 1 spatial struct. & 81.5 & 83.9 \\
        % 2 spatial struct. & 82.3 & 84.8 \\
        % 4 spatial struct. & 83.0 & 85.4 \\
        % 8 spatial struct. & 83.0 & 85.7 \\
        % 16 spatial struct. & 83.4 & 85.8 \\
        ConvPoint (16) & 83.4 & 85.8\\
        \textit{rank} & \textit{4} & \textit{4}\\

    \end{tabular}
    \end{minipage}
    \begin{minipage}{0.30\linewidth}
        \includegraphics[width=\linewidth]{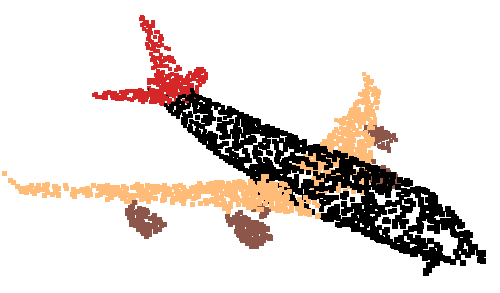}
        \includegraphics[width=\linewidth]{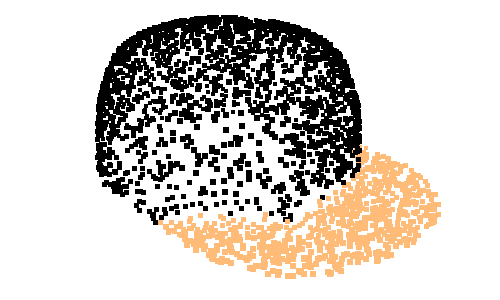}
        \includegraphics[width=\linewidth]{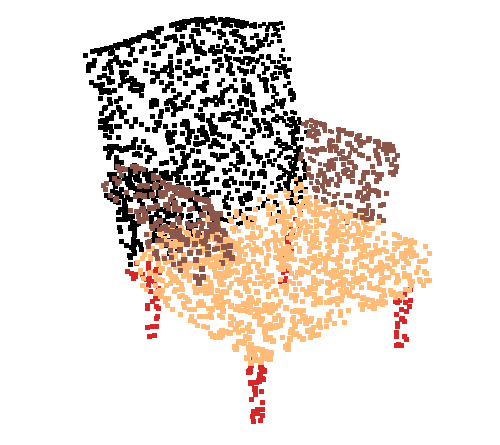}
        \includegraphics[width=\linewidth]{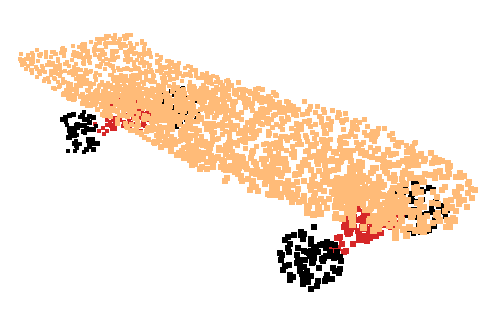}
    \end{minipage}
  \end{table}

  \subsubsection{Semantic segmentation}

  We now consider semantic segmentation for large-scale point clouds.
  As opposed to part segmentation, point clouds are now sampled in multi-object scenes outdoors and indoors.
  
  \paragraph*{Datasets}

  We use three datasets, one with indoor scenes, two with outdoors acquisitions.

  The \textit{Stanford 2D-3D-Semantics} dataset~\cite{2017arXiv170201105A} (S3DIS) is an indoor point cloud dataset for semantic segmentation.
  It is composed of six scenes, each corresponding to an office building floor.
  The points are labeled according to 13 classes: 6 building elements classes (floor, ceiling...), 6 office equipment classes (tables, chairs...) and a "stuff" class regrouping all the small equipment (computers, screens...)\ and rare items.
  For each scene considered as a test set, we train on the other five.

  The \textit{Semantic8}~\cite{sem3d} outdoor dataset is composed of 30 ground lidar scenes: 15 for training and 15 for evaluation.
  Each scene is generated from one lidar acquisition and the point cloud size ranges from 16 million to 430 million points.
  8 classes are manually annotated.

  Finally, the \textit{Paris-Lille 3D dataset (NPM3D)}~\cite{roynard2018paris} has been acquired using a Mobile Laser System. The training set contains four scenes taken from the two cites, totalizing 38 million points. The 3 tests scenes, with 10 million points each, were acquired on two other cities. The annotations correspond to 10 coarse classes from buildings to pedestrian.

  For both Semantic8 and NPM3D, test labels are unknown and evaluated on an online server.

  \paragraph*{Learning and prediction strategies}

  As the scenes may be large, up to hundred of millions of points, the whole point clouds cannot be directly given as input to the network.

\begin{figure}
    \centering
    \includegraphics[width=\linewidth]{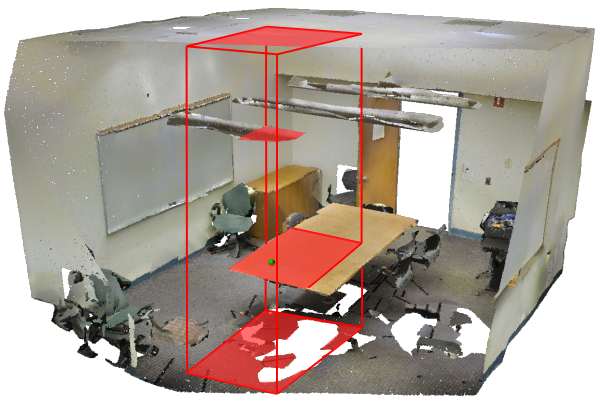}
    \caption{Point set selection for semantic segmentation. A point is selected in the original point cloud (green point) and an infinite vertical column centered around this point is defined (red frame) whose points are used for the network input (red points).}
    \label{fig:column}
  \end{figure}

  At training time, we randomly select points in the considered point cloud, and extract all the points in an infinite vertical column centered on this point, the column section is 2 meters wide for indoor scenes and 8 meters wide for outdoor scenes (Fig.~\ref{fig:column}).

  During testing, we compute a 2D occupancy pixel map with "pixel" size 0.1 meters for indoor scenes and 0.5 meters for outdoor scenes by projecting vertically on the horizontal plane. Then, we considered each occupied cell as a center for a column (same size as for training).
  
%   During testing, for a more systematic sampling of the space, we compute a 2D occupancy pixel map with pixel size 0.1 meters for indoor scenes and 0.5 meters for outdoor scenes. Then, we considered each occupied cell as a center for a column (same size as for training).

  For each column, we randomly select 8192 points which we feed as input to the network.
  Finally, the output scores are aggregated (summed) at point level and points not seen by the network receive the label of their nearest neighbor.

  \paragraph*{Improving efficiency with fusion of two networks learned on different modalities}

  First, the objective is to evaluate the influence of the color information as an input feature.

  We trained two networks, one with color information (\textit{RGB}) and one without (\textit{NoColor}).
    Let $P=\{(\mathbf{p},\mathbf{x})\}$ be the input point cloud.
    In the first case, \textit{RGB}, the input features is the color information, i.e., $\{\mathbf{x}\}=\{(r,g,b)\}$, $r,g,b$ being the red, green and blue values associated with each point.
    For the \textit{NoColor} model, input features do not hold color information: $\{\mathbf{x}\}=\{1\}$.

  Experimental results are presented in table~\ref{tab:_sem_seg}.
  We consider the line \textit{ConvPoint - RGB} and \textit{ConvPoint - NoColor} of table~\ref{tab:_sem_seg}(a) (S3DIS) and \ref{tab:_sem_seg}(b)(Semantic8).
  
  Intuitively, the \textit{RGB} network should outperform the other model on all categories: the \textit{RGB} point cloud is the \textit{NoColor} point cloud with additional color information.
  
  %As a matter of fact, using the geometry only is an information restriction with respect to the RGB model.
  However, according to the scores, the intuition is only partly verified.
  As an example in table~\ref{tab:_sem_seg}(a), the \textit{NoColor} model is much more efficient than the \textit{RGB} model on the \textit{column} class, mainly due to color confusion as walls and columns have most of the time the same color.
    To our understanding, when color is provided, the stochastic optimization process follows the easiest way to reach a training optimum, thus, giving too much importance to color information.
    In contrast, the \textit{NoColor} generates different features, based only on the geometry, and is more discriminating on some classes.
  %In practice, the geometry only model, as it cannot use the color, generates different features, these being more discriminative for some classes, which are not the same as the RGB model.
  
  Now looking at table~\ref{tab:_sem_seg}(b), we observe similar performances for both models.
  As these models use different inputs, they likely focus on different features of the scene.
  It should thus be possible to exploit this complementarity to improve the scores.

  \paragraph*{Model fusion}

  To show the interest of fusing models, we chose to use a residual fusion module~\cite{audebert_semantic_2016, boulch2017snapnet}.
  This approach have proven to produce good results for networks with different input modalities.
  Moreover, one advantage is that the two segmentation networks are first trained separately, the fusion module being trained afterward.
  This training process makes it possible to first, reuse the geometry only model if the RGB information is not available and second, to train with limited resources (see implementation details in section~\ref{sec_implementation}).

\begin{figure}
    \centering
    \includegraphics[width=0.9\linewidth]{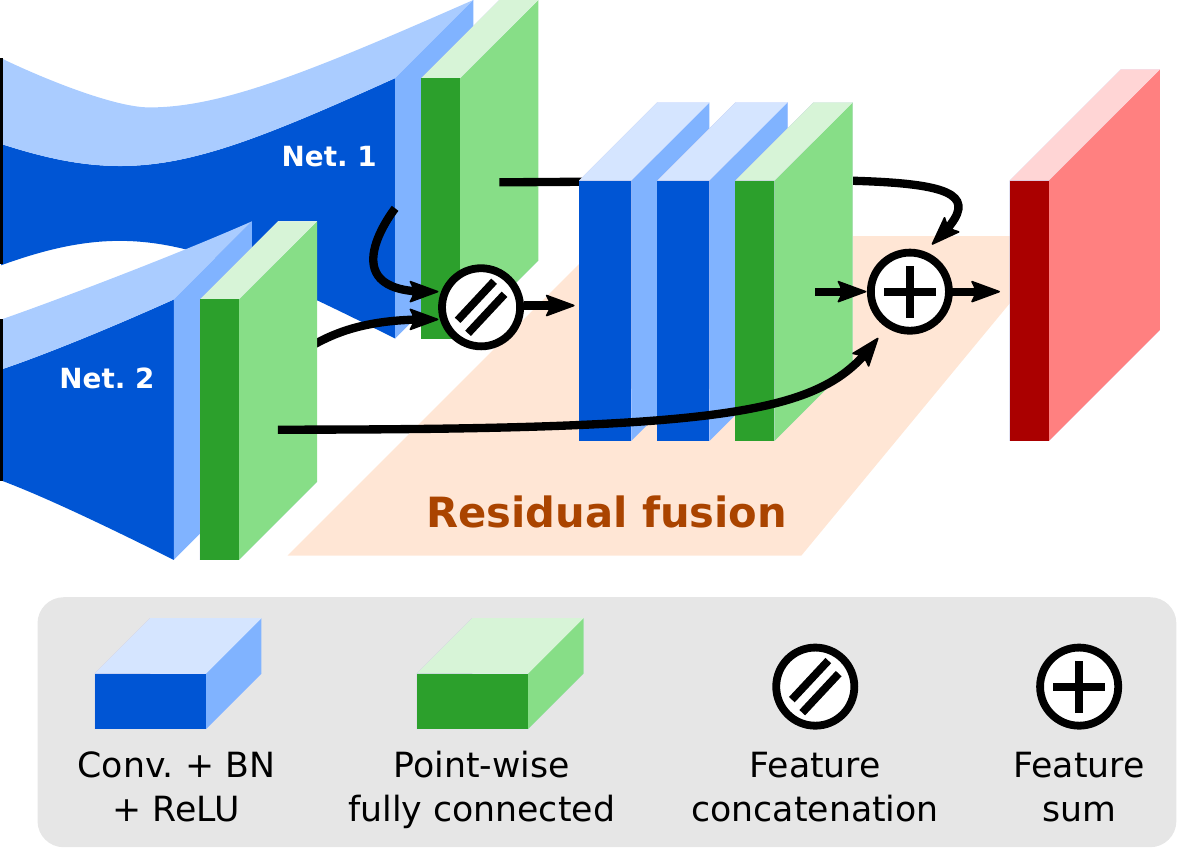}
    \caption{Segmentation networks: residual fusion.
        The input of the fusion module is the concatenation of the features collected before the fully-connected layer in networks 1 and 2. Its output is added as a correction to the sum of the outputs of networks 1 and 2.
    }
    \label{fig:net_fus}
  \end{figure}

  The fusion module is presented in Fig.~\ref{fig:net_fus}.
  The features before the fully-connected layer are concatenated, becoming the input of two convolutions and a point-wise linear layer.
  The outputs of both segmentation networks and the residual module are then added to form the final output.
  At training time, we add a dropout layer between the last convolution and the linear layer.

  In table~\ref{tab:_sem_seg}(a) and (b), the results of segmentation with fusion is reported at line \textit{ConvPoint - fusion}.
  As expected, the fusion increases the segmentation scores with respect to \textit{RGB} and \textit{NoColor} alone.

  On the S3DIS dataset (table~\ref{tab:_sem_seg}(a)), the fusion module obtains a better score on 10 out 13 categories and the average intersection over union is increased by 3.5\%.
  It is also interesting to note that, on categories for which the fusion is not better than with one single modality or the other, the score is close to the best mono-mode model.
  In other words, the fusion often improves the performance and in any case does not degrade it.

  On the Semantic8 dataset (table~\ref{tab:_sem_seg}(b)), we observe the same behavior.
  The gain is particularly high on the artefact class, which is one of the most difficult class: both mono-mode models reach 43-44\% while the fusion model reaches 61\%.
  It validates the fact that both \textit{RGB} and \textit{NoColor} models learn different features and that the fusion module is not only a sum of activations, but can select the best of both modalities.

  For comparison with official benchmarks, we use the fusion model.

%   \begin{table*}[]
%     \centering
%     \caption{S3DIS: testing set scene 2.}
%     \label{tab_s3dis_scene2}
%     \small
%     \begin{tabular}{c|c@{\hspace{0.2cm}}c|c@{\hspace{0.2cm}}c@{\hspace{0.2cm}}c@{\hspace{0.2cm}}c@{\hspace{0.2cm}}c@{\hspace{0.2cm}}c@{\hspace{0.2cm}}c@{\hspace{0.2cm}}c@{\hspace{0.2cm}}c@{\hspace{0.2cm}}c@{\hspace{0.2cm}}c@{\hspace{0.2cm}}c@{\hspace{0.2cm}}c}
%     Variant & OA& mIoU & ceiling& floor & wall& beam& column& window& door& table& chair& sofa& bookcase& board& clutter \\
%     \hline

% RGB         & \textit{85.65} & 53.07 & \textit{88.67} & \textit{96.86} & \textit{80.62} & 30.65 & 27.63 & \textit{58.53} & \textbf{62.73} & 43.37 & \textit{67.06} & 16.07 & \textit{45.73} & \textbf{21.79} & \textbf{50.22} \\
% Points only & 82.72 & \textit{55.12} & 80.94 & 96.67 & 76.16 & \textit{38.63} & \textbf{53.72} & 50.67 & 38.00 & \textbf{60.24} & \textbf{72.69} & \textbf{60.17} & 44.34 & 03.32 & 40.99 \\
% Fusion      & \textbf{86.01} & \textbf{58.10} & \textbf{89.33} & \textbf{96.94} & \textbf{81.27} & \textbf{38.69} & \textit{47.06} & \textbf{70.22} & \textit{62.15} & \textit{56.22} & 60.24 & \textit{40.49} & \textbf{50.49} & \textit{12.54} & \textit{49.72} \\
%       \end{tabular}
% \end{table*}

  % \begin{figure*}
  %   \centering
  %   \includegraphics[width=0.8\linewidth]{images/net_fus}
  %   \caption{Segmentation networks: residual fusion.}
  %   \label{fig:net_fus}
  % \end{figure*}

  \paragraph*{Large-scale datasets: comparison with the state of the art}

  Table~\ref{tab:_sem_seg} presents also, for comparison, the scores obtained by other methods in the literature.
  Our approach is competitive with the state of the art on the three datasets.

  On S3DIS, we place second behind KPConv~\cite{thomas2019KPConv} in term of average intersection over union (mIoU), while being first on several categories.
  It can be noted that approaches sharing concepts with ours, such as PCCN~\cite{wang2018deep} or PointCNN~\cite{NIPS2018_7362}, do not perform as well as ours.

  On Semantic8, we report the state of the benchmark leaderboard at the time of article writing (for entries that are not anonymous).
  PointNet++ has two entries in the benchmark, we only reported the best one.
  Our convolutional network for segmentation places first before Superpoint Graph (SPG)~\cite{landrieu2018large} and SnapNet~\cite{boulch2017snapnet}.
  It differs greatly from SPG, which relies on a pre-segmentation of the point cloud, and SnapNet, which uses a 2D segmentation network on virtual pictures of the scene to produce segments.
  We surpass the PointNet++ by 13\% on the average IoU.
  We perform particularly well on car and artifacts detection where other methods, except for SPG get relatively low results.

  Finally on NPM3D Paris-Lille dataset (table~\ref{tab:_sem_seg}(c)), we also report the official benchmark at the time the paper was written.
  Based on the average IoU, we place second surpassed only by KPConv~\cite{thomas2019KPConv}.
  Our approach is the best or second best for 6 out of 9 categories.
  The second place is explained mostly by the relatively low score on pedestrian and trash cans.
  These are particularly difficult classes, due to their variability and the low number of instances in the train set.
 
% \begin{figure}[t]
%   \centering
%   \begin{tabular}{c}
%   \includegraphics[width=0.8\linewidth]{images/influence}\\
%     (a) Weighting functions of some kernel elements \\
%   \includegraphics[width=0.8\linewidth]{images/filters}\\
%     (b) Filters
%   \end{tabular}
%   \caption{Weighting function and filters.}
%   \label{fig_filters}
% \end{figure}

\begin{figure}[t]
  \centering
  \begin{tabular}{c|c}
  \includegraphics[width=0.4\linewidth]{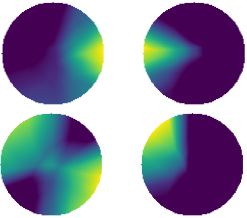} & \includegraphics[width=0.4\linewidth]{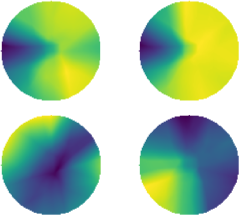}\\
    (a) Weighting functions &
    (b) Filters
  \end{tabular}
  \caption{Weighting function and filters for the first convolution layer of the classification model trained of MNIST.}
  \label{fig_filters}
\end{figure}

\begin{table}[]
    \centering
    \caption{Influence of the spatial sampling number, ShapeNet dataset.}
    \label{tab:samplings}
    \begin{tabular}{c|ccccc}
        Spatial samplings   & 1 & 2 & 4 & 8 & 16 \\
        \hline
        mIoU                & 83.9 & 84.8 & 85.4 & 85.7 & 85.8 \\
    \end{tabular}
\end{table}

\begin{table*}[t]
  \caption{Large scene semantic segmentation benchmarks.}
  \label{tab:_sem_seg}
  \centering
  \small
  
   \colorbox{Green}{\textbf{Best score}}, \colorbox{YellowGreen}{\textit{second best}} and \colorbox{GreenYellow}{third best}.
  
  \vspace{0.1cm}

  (a) S3DIS 

%   \begin{tabular}{c|Mc@{\hspace{0.1cm}}c@{\hspace{0.1cm}}c|c@{\hspace{0.1cm}}c@{\hspace{0.1cm}}c@{\hspace{0.1cm}}c@{\hspace{0.1cm}}c@{\hspace{0.1cm}}c@{\hspace{0.1cm}}c@{\hspace{0.1cm}}c@{\hspace{0.1cm}}c@{\hspace{0.1cm}}c@{\hspace{0.1cm}}c@{\hspace{0.1cm}}c@{\hspace{0.1cm}}c}
  \begin{tabular}{c|MMM|MMMMMMMMMMMMM}
    Method & OA & mAcc & mIoU & ceil. & floor & wall & beam & col. & wind. & door & chair & table & book. & sofa & board & clut. \\
    \hline
    % PointNet~\cite{qi2017pointnet} &78.5 & 66.2& 47.6& 88.0& 88.7& 69.3& 42.4& 23.1& 47.5& 51.6& 54.1& 42.0& 9.6& 38.2& 29.4& 35.2 \\
    % SPGraph~\cite{landrieu2018large} &85.5& 73.0& 62.1& 89.9& 95.1& 76.4& 62.8& 47.1& 55.3& 68.4 & \textbf{73.5}& 69.2& \textbf{63.2}& 45.9& 8.7& 52.9 \\
    % RSNet~\cite{huang2018recurrent}&  -&  66.45&  56.47&  92.48&  92.83&  78.56&  32.75&  34.37&  51.62&  68.11&  60.13&  59.72&  50.22&  16.42&  44.85&  52.03 \\
    % PCCN~\cite{wang2018deep}& - & 67.01 & 58.27& 92.26& 96.20& 75.89&  0.27& 5.98& \textbf{69.49}& 63.45& 66.87& 65.63& 47.28& 68.91&  \textbf{59.10}&  46.22 \\
    % PointCNN~\cite{NIPS2018_7362}&   88.14 &  \textbf{75.61}&  65.39 &  \textbf{94.78}&  97.3 &   75.82&  \textbf{63.25}&  \textbf{51.71}&  58.38&  57.18&  71.63&  69.12&  39.08&  61.15 &  52.19 & 58.59 \\

    % KPConv~\cite{thomas2019KPConv} & - & 79.1 & 70.6 & 93.6 & 92.4 & \textbf{83.1} & \textbf{63.9} & \textbf{54.3} & 66.1 & \textbf{76.6} & 64.0 & 57.8 & \textbf{69.3} & \textbf{74.9} & \textbf{61.3} & 60.3 \\
    % ConvPoint - Fusion              & \textbf{88.36} & - & \textbf{66.71} &  94.50 & \textbf{97.49} & 80.45 & 45.33 & 32.40 &  63.61 & 70.08 & 70.47 & \textbf{77.32} & 53.98 & \textbf63.10 & 54.55 & \textbf{63.89}\\
    Pointnet~\cite{qi2017pointnet}  & 78.5 & 66.2 & 47.6 & 88.0 & 88.7 & 69.3 & 42.4 & 23.1 & 47.5 & 51.6 & 42.0 & 54.1 & 38.2 &  9.6 & 29.4 & 35.2 \\
    RSNet~\cite{huang2018recurrent} &    - & 66.5 & 56.5 & 92.5 & 92.8 & 78.6 & 32.8 & 34.4 & 51.6 & 68.1 & 60.1 & 59.7 & 50.2 & 16.4 & 44.9 & 52.0 \\
    PCCN~\cite{wang2018deep}        &    - & 67.0 & 58.3&  92.3 & 96.2 & 75.9 & 0.27 &  6.0 & \cellcolor{Green}\textbf{69.5} & 63.5 & 65.6 & 66.9 & \cellcolor{YellowGreen}\textit{68.9} & 47.3 & \cellcolor{YellowGreen}\textit{59.1} & 46.2 \\
    SPGraph~\cite{landrieu2018large}& 85.5 & \cellcolor{GreenYellow}73.0 & 62.1 & 89.9 & 95.1 & 76.4 & \cellcolor{GreenYellow}62.8 & \cellcolor{GreenYellow}47.1 & 55.3 & \cellcolor{GreenYellow}68.4 & \cellcolor{GreenYellow}73.5 & \cellcolor{GreenYellow}69.2 & 63.2 & 45.9 &  8.7 & 52.9 \\ 
    PointCNN~\cite{NIPS2018_7362}   & \cellcolor{YellowGreen}\textit{88.1} & \cellcolor{YellowGreen} \textit{75.6} & \cellcolor{GreenYellow}65.4 & \cellcolor{GreenYellow} 94.8 & \cellcolor{YellowGreen}\textit{97.3} & 75.8 & \cellcolor{YellowGreen}\textit{63.3} & \cellcolor{YellowGreen}\textit{51.7} & 58.4 & 57.2 & 71.6 & 69.1 & 39.1 & \cellcolor{YellowGreen}\textit{61.2} & 52.2 & 58.6 \\
    KPConv~\cite{thomas2019KPConv}  &    - & \cellcolor{Green}\textbf{79.1} & \cellcolor{Green}\textbf{70.6} & 93.6 & 92.4 & \cellcolor{Green}\textbf{83.1} & \cellcolor{Green}\textbf{63.9} & \cellcolor{Green}\textbf{54.3} & \cellcolor{YellowGreen}\textit{66.1} & \cellcolor{Green}\textbf{76.6} & 57.8 & 64.0 & \cellcolor{Green}\textbf{69.3} & \cellcolor{Green}\textbf{74.9} & \cellcolor{Green}\textbf{61.3} & \cellcolor{GreenYellow}60.3 \\
    \hline
    % ConvPoint - Fusion              & \textbf{88.4} &    - & \textit{66.7} & \textit{94.5} & \textbf{97.5} & \textit{80.5} & 45.3 & 32.4 & 63.6 & \textbf{70.1} & \textbf{77.3} & \textbf{70.5} & 63.1 & 54.0 & 54.6 & \textbf{63.9} \\
    ConvPoint - RGB & \cellcolor{GreenYellow}87.9 & - & 64.7 & \cellcolor{Green}\textbf{95.1} & \cellcolor{Green}\textbf{97.7} & \cellcolor{GreenYellow}80.0 & 44.7 & 17.7 & 62.9 & 67.8 & \cellcolor{YellowGreen}\textit{74.5} & \cellcolor{YellowGreen}\textit{70.5} & 61.0 & 47.6 & 57.3 & \cellcolor{YellowGreen}\textit{63.5} \\
    ConvPoint - NoColor & 85.2 & - & 62.6 & 92.8 & 94.2 & 76.7 & 43.0 & 43.8 & 51.2 & 63.1 & 71.0 & 68.9 & 61.3 & 56.7 & 36.8 & 54.7 \\
    ConvPoint - Fusion & \cellcolor{Green}\textbf{88.8} & - & \cellcolor{YellowGreen}\textit{68.2} & \cellcolor{YellowGreen}\textit{95.0} & \cellcolor{YellowGreen}\textit{97.3} & \cellcolor{YellowGreen}\textit{81.7} & 47.1 & 34.6 & \cellcolor{GreenYellow}63.2 &\cellcolor{YellowGreen} \textit{73.2} &\cellcolor{Green} \textbf{75.3} &\cellcolor{Green} \textbf{71.8} & \cellcolor{GreenYellow}64.9 & \cellcolor{GreenYellow}59.2 & \cellcolor{GreenYellow}57.6 &\cellcolor{Green} \textbf{65.0} \\
    % Fusion & \textbf{88.8} & - & 68.2 & \textbf{95.0} & \textbf{97.3} & \textit{81.7} & 47.1 & 34.6 & 63.0 & \textit{73.2} & \textbf{75.3} & \textbf{71.8} & 64.9 & 59.2 & 57.6 & \textbf{65.0} \\
  \end{tabular}
    \\~\\

  (b) Semantic8 

%   \begin{tabular}{c|c@{\hspace{0.2cm}}c|c@{\hspace{0.2cm}}c@{\hspace{0.2cm}}c@{\hspace{0.2cm}}c@{\hspace{0.2cm}}c@{\hspace{0.2cm}}c@{\hspace{0.2cm}}c@{\hspace{0.2cm}}c}
  \begin{tabular}{c|MM|MMMMMMMM}
      Method & mIoU & OA & Man made & Natural & High veg. & Low veg. & Buildings & Hard scape & Artefacts & Cars\\
      \hline
      TML-PC~\cite{montoya2014mind}& 	0.391&	0.745&	0.804&	0.661&	0.423&	0.412&	0.647&	0.124&	0.000&	0.058 \\
      TMLC-MS~\cite{hackel2016fast}&	0.494&	0.850&	0.911&	0.695&	0.328&	0.216&	0.876&	0.259&	0.113&	0.553 \\
      PointNet++~\cite{qi2017pointnet++}&	0.631& 0.857& 0.819& 0.781&	0.643&	0.517&	0.759&	0.364&	0.437&	0.726 \\
      SnapNet~\cite{boulch2017snapnet}&    0.674&	0.910&	0.896&	\cellcolor{GreenYellow}0.795 &	0.748&	0.561&	0.909&	0.365&	0.343&	0.772 \\
      SPGraph~\cite{landrieu2018large} &	\cellcolor{YellowGreen}\textit{0.762} &	\cellcolor{GreenYellow}0.929 &	0.915 &	0.756&	\cellcolor{Green}\textbf{0.783}&	\cellcolor{YellowGreen}\textit{0.717} &	0.944 &	\cellcolor{Green}\textbf{0.568}&	\cellcolor{YellowGreen}\textit{0.529} &	\cellcolor{Green}\textbf{0.884} \\
      \hline

      ConvPoint - RGB & \cellcolor{GreenYellow}0.750 & \cellcolor{Green}\textbf{0.938} & \cellcolor{Green}\textbf{0.934} & \cellcolor{Green}\textbf{0.847} & \cellcolor{GreenYellow}0.758 & \cellcolor{GreenYellow}0.706 & \cellcolor{GreenYellow}0.950 & \cellcolor{YellowGreen}\textit{0.474} & 0.432 & \cellcolor{YellowGreen}\textit{0.902} \\
      ConvPoint - NoColor    & 0.726 & 0.927 & \cellcolor{GreenYellow}0.918 & 0.788 & 0.748 & 0.646 & \cellcolor{Green}\textbf{0.962} & 0.451 & \cellcolor{GreenYellow}0.442 & 0.856 \\

      ConvPoint - Fusion & \cellcolor{Green}\textbf{0.765} & \cellcolor{YellowGreen}\textit{0.934} & \cellcolor{YellowGreen}\textit{0.921} & \cellcolor{YellowGreen}\textit{0.806} & \cellcolor{YellowGreen}\textit{0.760} & \cellcolor{Green}\textbf{0.719} & \cellcolor{YellowGreen}\textit{0.956} & \cellcolor{GreenYellow}0.473 & \cellcolor{Green}\textbf{0.611} & \cellcolor{GreenYellow}0.877\\
  \end{tabular}\\~\\
  
  (c) NPM3D

%   \begin{tabular}{c|c|ccccccccc}
  \begin{tabular}{c|c|MMMMMMMMM}
    Name & mIoU & Ground & Building & Pole & Bollard & Trash can & Barrier & Pedestrian & Car & Natural \\
    \hline
    MSRRNetCRF & 65.8 & 99.0 & \cellcolor{Green}\textbf{98.2} & 45.8 & 15.5 & \cellcolor{GreenYellow}64.8 & \cellcolor{Green}\textbf{54.8} & 29.9 & \cellcolor{Green}\textbf{95.0} & \cellcolor{GreenYellow}89.5\\
    EdConvE    & 66.4 & 99.2 & 91.0 & 41.3 & 50.8 & \cellcolor{YellowGreen}\textit{65.9} & 38.2 & 49.9 & 77.8 & 83.8\\
    RFMSSF     & 56.3 & 99.3 & 88.6 & 47.8 & 67.3 &  2.3 & 27.1 & 20.6 & 74.8 & 78.8\\
    MS3DVS     & 66.9 & 99.0 & \cellcolor{GreenYellow}94.8 & 52.4 & 38.1 & 36.0 & \cellcolor{GreenYellow}49.3 & \cellcolor{GreenYellow}52.6 & \cellcolor{GreenYellow}91.3 & 88.6\\
    HDGCN      & \cellcolor{GreenYellow}68.3 & \cellcolor{GreenYellow}99.4 & 93.0 & \cellcolor{GreenYellow}67.7 & \cellcolor{GreenYellow}75.7 & 25.7 & 44.7 & 37.1 & 81.9 & \cellcolor{YellowGreen}\textit{89.6}\\
    %KP-FCNN~\cite{thomas2019KPConv}	& \textbf{75.9} & \textit{99.5} & 93.2 & \textbf{69.3} & \textit{82.2} & 48.8 & 44.3 & \textbf{62.0} & \textit{93.6} & \textbf{90.4}\\
    KP-FCNN~\cite{thomas2019KPConv} &	\cellcolor{Green}\textbf{82.0} & \cellcolor{Green}\textbf{99.5} & 94.0 & \cellcolor{YellowGreen}\textit{71.3} & \cellcolor{YellowGreen}\textit{83.1} & \cellcolor{Green}\textbf{78.7} & 47.7 & \cellcolor{Green}\textbf{78.2} & \cellcolor{YellowGreen}\textit{94.4} & \cellcolor{Green}\textbf{91.4}\\
    \hline
    % ConvPoint - Fusion & \textit{75.2}	& \textbf{99.6} & \textit{96.8} & \textbf{69.3} & \textbf{86.3} & 54.3 & \textit{53.7} & 35.3 & 91.1 & \textbf{90.4}\\
    ConvPoint - Fusion & \cellcolor{YellowGreen}\textit{75.9} &	\cellcolor{Green}\textbf{99.5} &	\cellcolor{YellowGreen}\textit{95.1} &	\cellcolor{Green}\textbf{71.6} &	\cellcolor{Green}\textbf{88.7} &	46.7 &	\cellcolor{YellowGreen}\textit{52.9} &	\cellcolor{YellowGreen}\textit{53.5} &	89.4 &	85.4
  \end{tabular}

\end{table*}

% \begin{table*}[]
%   \caption{NPM3D.}
%   \label{tab:npm3d}
%   \centering
%   \small

%   \begin{tabular}{c|cccccccccc}
%   Name & Av. IoU & Ground & Building & Pole & Bollard & Trash can & Barrier & Pedestrian & Car & Natural \\
%   \hline
%   MSRRNetCRF & 65.8 & 99.0 & \textbf{98.2} & 45.8 & 15.5 & 64.8 & \textbf{54.8} & 29.9 & \textbf{95.0} & 89.5\\
%   EdConvE    & 66.4 & 99.2 & 91.0 & 41.3 & 50.8 & \textbf{65.9} & 38.2 & 49.9 & 77.8 & 83.8\\
%   RFMSSF     & 56.3 & 99.3 & 88.6 & 47.8 & 67.3 &  2.3 & 27.1 & 20.6 & 74.8 & 78.8\\
%   MS3DVS     & 66.9 & 99.0 & 94.8 & 52.4 & 38.1 & 36.0 & 49.3 & 52.6 & 91.3 & 88.6\\
%   HDGCN      & 68.3 & 99.4 & 93.0 & 67.7 & 75.7 & 25.7 & 44.7 & 37.1 & 81.9 & 89.6\\
%   KP-FCNN	   & \textbf{75.9} & 99.5 & 93.2 & 69.3 & 82.2 & 48.8 & 44.3 & 62.0 & 93.6 & \textbf{90.4}\\
%   \hline
%   ConvPoint - Fusion & 75.2	& \textbf{99.6} & 96.8 & \textbf{69.3} & \textbf{86.3} & 54.3 & 53.7 & 35.3 & 91.1 & \textbf{90.4}
%   \end{tabular}

% \end{table*}

\subsection{Emprical properties of the convolutional layer.}

\paragraph*{Filter visualization}
Fig.~\ref{fig_filters} presents a visualization of some characteristics of the first convolutional layer of the 2D classification model trained on MNIST.

Fig.~\ref{fig_filters}(a) shows the weighting function associated with four of the sixty-four kernel elements of the first convolutional layer.

% The first line (Fig.~\ref{fig_filters} (a)) shows the weighting function associated with four of the sixty-four kernel elements.
These weights are the output of the MLP function.
As expected, their nature varies a lot depending on the kernel element, underlying different regions of interest for each kernel element.

Fig.~\ref{fig_filters}(b) shows the resulting convolutional filters.
These are computed using the previously presented weighting functions, multiplied by the weights of each kernel element ($\mathbf{w}$) and summed over the kernel elements.
As with discrete CNNs, we observe that the network has learned various filters, with different orientations and shapes.

\subsubsection{Influence of random selection of output points}
\label{sec:spatial_sampling}

The strategy for selecting points of the output $\{\mathbf{q}\}$ is stochastic at each layer, i.e., for a fixed input, two runs through the same layer may lead to two different $\{\mathbf{q}\}$'s.
Therefore, the outputs $\{\mathbf{y}\}$'s may also be different, and so may be the predicted labels.
To increase robustness, we aggregate several outputs computed with the same network and the same input point cloud.
This is referred to as the number of sampling (from 1 to 16) in table~\ref{tab:samplings}.
We observe an improvement of the performances with the number of spatial sampling.
In practice, we only use up to 16 samplings because a larger number does not significantly improve the scores.

This procedure shares similarities with the approaches used for image processing for test set augmentation such as image crops~\cite{simonyan2014very, szegedy2016rethinking}.
The main difference resides in the fact that the output variation is not a result of input variation but is inherent to the network, i.e., the network is not deterministic.

\subsubsection{Robustness to point cloud size and neighborhood size}

\begin{figure}
  \centering
  \includegraphics[width=\linewidth]{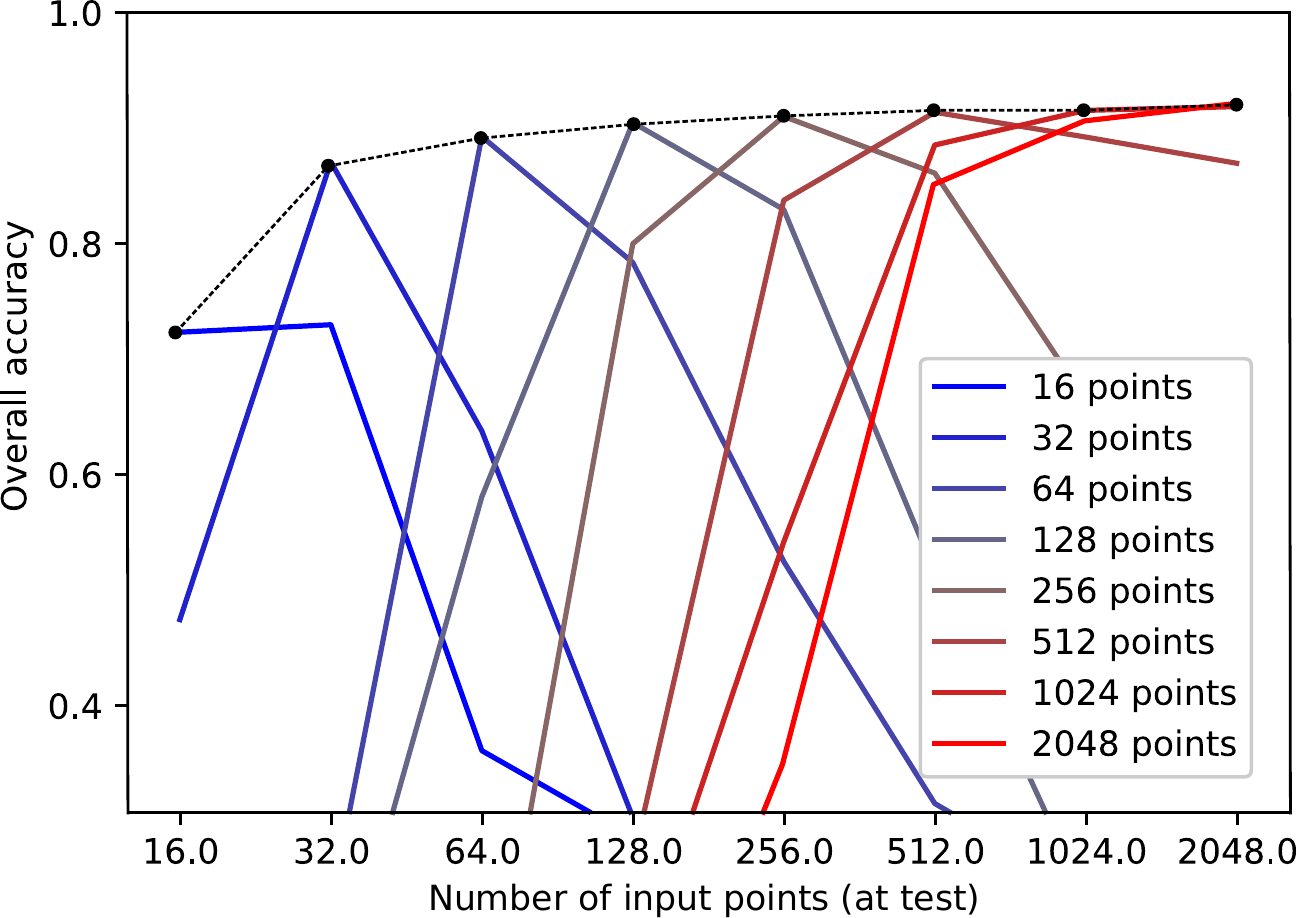}
  \caption{Influence of a varying number of input points at training time versus at test time on ModelNet40. 
  Each colored curve shows the performance of a function of  model trained with a given input point cloud size when tested on different numbers of input points.
  }
  \label{fig:npts}
\end{figure}

In order to evaluate the robustness to test conditions that are different from the training ones, we propose two experiments.
As stated in section~\ref{sec_conv}, the definition of the convolutional layer does not require a fixed input size.
However for a gain in performance and time, we trained the networks with minibatches, fixing the input size at training.
We evaluate the influence of the input size at inference on the performance of a model trained with fix input size.
Results are presented in Fig.~\ref{fig:npts} for the ModelNet40 classification dataset.
Each curve (from blue to red) is an instance of the classification network, trained with 16, 32, \dots, 2048 input points.
The black dots are the scores for each model at their native input size.
The dashed curve describe a theoretical model performing as well as the best model for each input size.
Please note that the horizontal scale is a log scale and that each step is doubling the number of input points.
A first observation is that almost each model performs the best at its native input size and that very few points are needed on ModelNet40 to reach decent performances: with 32 points, the performance already reaches to 85\%.
Besides, the larger the training size is, the more robust to size variation the model becomes.
While the model trained on 32 points see its performance drop by 25\% with $\pm$50\% points,
the model trained with 2048 points still reaches 82\% (a drop of 10\%) with only 512 points (4 times less).

Second, in our formulation, the neighborhood size $|X|$ for each convolution layer remains a variable parameter after training,
i.e., it is possible to change the  neighborhood size at test time.
It is the reason why, in equation~\eqref{eq:conv}, the normalizing weight $1/|X|$ (average with respect to input size) has been added to be robust to neighborhood size variation.
In addition to the robustness provided by averaging, the variation of $k$ somehow simulates a density variation of the points for the layer.
We evaluate the robustness to such variations in table~\ref{tab:kvariation}, on the classification dataset ModelNet40, with a single model trained with the default configuration (see Fig.~\ref{fig:net_classif}).
We report the impact of $k$ on the first layer (table~\ref{tab:kvariation}(a)) and on the fourth layer (table~\ref{tab:kvariation}(b)).
As expected, even though the best score is reached for the default $k$, the layer is robust to a high variation of $k$ values: the overall accuracy loss is lower than 2\% when $k$ is 2 times larger or smaller.
A particularly interesting feature is that the first layer, which extract local features, is more robust to a decreasing $k$, making the features more local than a increasing $k$.
It is the opposite for the fourth layer: global features are more robust to an increase of $k$ than a decreasing $k$.

\begin{table}
\caption{ModelNet40 classification scores with a variation of neighborhood sizes $k$ at test time.}
\label{tab:kvariation}

\centering
  \vspace{0.5em}

\begin{tabular}{cccccc}
    (a) &\multicolumn{5}{l}{\textit{First layer (default $k=32$)}}\\
    & /2 & /1.5 & $\times$1 & $\times$1.5 & $\times$2 \\
     & 91.0 & 92.2 & 92.5 & 91.6 & 90.2 \\
     \hline
     (b) & \multicolumn{5}{l}{\textit{4$^{th}$ layer (default $k=16$)}}\\
    & /2 & /1.5 & $\times$1 & $\times$1.5 & $\times$2 \\
     & 90.9 & 91.2 & 92.5 & 91.9 & 91.4\\
\end{tabular}

  \vspace{0.5em}
\textit{Note: scores are computed with a 16-element spatial structure. The model was trained with the default configuration.}
\end{table}

 \begin{table*}
  \caption{Training and inference timings.}
  \label{tab:timings}
  \vspace{0.5em}
  \begin{minipage}[t]{0.52\linewidth}
    (a) Comparison with PointCNN~\cite{NIPS2018_7362} on ModelNet40 and ShapeNet, computed with \textit{Config. 1}
    \begin{center}
    \begin{tabular}{c@{~}c@{~}c|c|@{~}c|c|@{~}c}
                  &             &             & \multicolumn{2}{c|}{PointCNN} & \multicolumn{2}{c}{ConvPoint} \\
      \cline{4-7}
      Dataset     & Point & Batch & Training  & Test  & Training  & Test \\
                  & num.  & size  & (epoch)   &       & (epoch)   &      \\
      \hline
      ModelNet40  & 1024        & 32          & 58 s       & 8 s& 42 s               & 7 s        \\
                  &             & 128         & 51 s       & 7 s& 35 s               & 7 s       \\
                  & 2048        & 32          & 80 s  & 11 s& 53 s              & 9 s        \\
                  &             & 128         & \multicolumn{2}{c|}{-} & 45 s               & 8 s        \\
      \hline
      ShapeNet    & 2048        & 4           & 1320 s     & 120 s & 255 s        & 41 s       \\
                  &             & 8           & \multicolumn{2}{c|}{-}& 185 s           & 37 s       \\
                  &             & 64          & \multicolumn{2}{c|}{-}& 116 s           & 29 s       \\
    \end{tabular}
    
    \end{center}
  \end{minipage}
  \hfill
  \begin{minipage}[t]{0.45\linewidth}
    (b) Comparison between large segmentation network and fusion architecture, computed on the S3DIS dataset. Timings are given in milliseconds.
    \begin{center}
      \begin{tabular}{c|cccccc}
        Batch size      & 1 & 2 & 4 & 8 & 16 & 32 \\
        \hline
        \multicolumn{2}{l}{\textit{Segmentation network}}\\
        Config. 1 &  93 & 69 & 46 & 40 & 39 & 38 \\
        Config. 2 & 155 & 95 & 89 & 81 &  - &  - \\
        \hline
        \multicolumn{2}{l}{\textit{Fusion architecture}} \\
        Config. 1 & 230 & 170 & 110 & 103 & 101 & - \\
        Config. 2 & 418 & 282 & 232 &   - &   - & - \\
      \end{tabular}
    \end{center}
  \end{minipage}

  \begin{center}
    \textit{Config. 1}: Middle-end configuration. Intel Xeon E5-1620 v3, 3.50 GHz, 8 cores, 8 threads + Nvidia GTX 1070 8Gb
    
    \textit{Config. 2}: Low-end configuration. Intel Core i7-5500U+, 2.40GHz, 2 cores, 4 threads + Nvidia GTX 960M 2Gb
    
    The symbol "-" corresponds to setups (batch size and number of points) that exceed GPU memory. 
  \end{center}
  
\end{table*}

\section{Computation times and implementation details}
\label{sec_implementation}

In our experiments, the convolutional pipeline was implemented using Pytorch~\cite{paszke2017automatic}.
The neighborhood computation are done with NanoFLANN~\cite{blanco2014nanoflann} and parallelized over CPU cores using OpenMP~\cite{Dagum:1998:OIA:615255.615542}.

Table~\ref{tab:timings} presents the computation times.
We consider two hardware configurations: a desktop workstation (Config.~1) and a low-end configuration, i.e., a gaming laptop (Config.~2). For instance, Config.~2 could correspond to a hardware specification for embedded devices.

Table~\ref{tab:timings}(a) is a comparison with PointCNN~\cite{NIPS2018_7362} which was reported as the fastest in \cite{NIPS2018_7362} among other methods.
We used the code version available one the official PointCNN repository at the time of submission, and used the recommended network architecture.
For both framework, we ran experiments on the \textit{Config.~1} computer.
On the ModelNet40 classification dataset, our model is about 30\% faster than PointCNN for training, but inference times are similar.
The difference is more significant on the ShapeNet segmentation dataset.
For a batch size of 4, our segmentation framework is more than 5 times faster for training, and 3 times faster at test time.

Moreover, we also show that our ConvPoint is more memory efficient than the implementation of PointCNN. The "-" symbol in table~\ref{tab:timings} indicates batch sizes / numbers of points configurations that exceed the GPU memory.
For example, we can train the classification model with 2048 points and batch size of 128, which is not possible with PointCNN (on an Nvidia GTX 1070 GPU). The same goes on ShapeNet where we can train with a batch size of 64 while PointCNN already uses too much GPU memory with batch size 8.

In table~\ref{tab:timings}(b), we report timings for the large-scale segmentation network and the fusion architecture, with a point cloud size of 8192. Note that for training this network we used NVidia GTX 1080Ti GPU.
These timings represent the inference time only (neighborhood computation and convolutions), not data loading.
We first observe that even the fusion architecture (two segmentation networks and a fusion module) can be run on the small configuration.
Second, as for CNN with images, we benefit from using batches which reduce per point cloud computation time.
Finally, our implementation is efficient given that for the segmentation network, we are able to process from 100,000 points (Config.~2) to 200,000 points (Config.~1) per second.

\section{Discussion and limitations}

\paragraph{Convolutional layer} First, our convolution design is agnostic to the object scales, due to neighborhood normalization to the unit ball.
It is of interest for non metric data such as CAD models or photogrammetric point clouds where scales are not always available.
On the contrary, in metric scans, object sizes are valuable information (e.g., humans have almost all the time similar sizes) but
removing the normalization would cause the kernel and the input points to have different volumes.

Second, an alternative is to use a fixed-radius neighborhood instead of a fix number of neighbors.
As pointed out in~\cite{thomas2019KPConv}, the resulting features would be more related to geometry and less to sampling.
However, the actual code optimization to speed up computation such as batch training would be inapplicable due to a variable number of neighbors in a batch.

\paragraph{Input features}
Another perspective is to explore the use of precomputed features as inputs.
In this study, we only use raw data for network inputs: RGB colors when available, or all features set to 1 otherwise.
In the future, we will work on feeding the networks with extra features such as normals or curvatures.

\paragraph{Network architecture}
Finally, we proposed two networks architectures that are widely inspired from computer vision models.
It would be interesting to explore further variations of network architectures.
As our formulation generalizes the discrete convolution, it is possible to transpose more CNN architectures, such as residual networks.

\section{Conclusion}

In this paper, we presented a new CNN framework for point cloud processing.
The proposed formulation is a generalization of the discrete convolution for sparse and unstructured data.
It is flexible and computationally efficient, which it makes it possible to build various network architectures for classification, part segmentation and large-scale semantic segmentation.
Through several experiments on various benchmarks, real and simulated, we have shown that our method is efficient and at state of the art.

%TODO
\small
\bibliographystyle{abbrv}

\bibliography{refs.bib}

\end{document}